\documentclass[9pt,shortpaper,twoside,web]{IEEEtran}
\usepackage{cite}
\usepackage{amsmath,amssymb,amsfonts}
\usepackage{graphicx}
\usepackage{textcomp}
\usepackage{afterpage}
\usepackage{tabularx} 
\usepackage{multirow}
\usepackage{float}
\usepackage[table,xcdraw]{xcolor}
\usepackage{hyperref}
\usepackage{algorithm}
\usepackage{algpseudocode}

\begin{document}
\title{A Self-Supervised Framework for Improved Generalisability in Ultrasound B-mode Image Segmentation}
\author{Edward Ellis, Andrew Bulpitt, Nasim Parsa, Michael F Byrne and Sharib Ali
\thanks{This work was supported by UK Research and Innovation (UKRI) [CDT grant number EP/S024336/1] and Satisfai Health.} 
\thanks{E. Ellis, A. Bulpitt and S. Ali are with the School of Computer Science, Faculty of Engineering and Physical Sciences, University of Leeds, LS2 9JT, Leeds, United Kingdom (emails: \{scee, a.j.bulpitt, s.s.ali\}@leeds.ac.uk)}
\thanks{N. Parsa and M. Byrne are with Satisfai Health, 1050 Homer Street, Suite 390, Vancouver, BC V6B 2W9, Canada}}

\maketitle
\begin{abstract}
Ultrasound (US) imaging is clinically invaluable due to its noninvasive and safe nature. However, interpreting US images is challenging, requires significant expertise, and time, and is often prone to errors. Deep learning offers assistive solutions such as segmentation. Supervised methods rely on large, high-quality, and consistently labeled datasets, which are challenging to curate. Moreover, these methods tend to underperform on out-of-distribution data, limiting their clinical utility. Self-supervised learning (SSL) has emerged as a promising alternative, leveraging unlabeled data to enhance model performance and generalisability. We introduce a contrastive SSL approach tailored for B-mode US images, incorporating a novel Relation Contrastive Loss (RCL). RCL encourages learning of distinct features by differentiating positive and negative sample pairs through a learnable metric. Additionally, we propose spatial and frequency-based augmentation strategies for the representation learning on US images.

Our approach significantly outperforms traditional supervised segmentation methods across three public breast US datasets, particularly in data-limited scenarios. Notable improvements on the Dice similarity metric include a 4\% increase on 20\% and 50\% of the BUSI dataset, nearly 6\% and 9\% improvements on 20\% and 50\% of the BrEaST dataset, and 6.4\% and 3.7\% improvements on 20\% and 50\% of the UDIAT dataset, respectively. Furthermore, we demonstrate superior generalisability on the out-of-distribution UDIAT dataset with performance boosts of 20.6\% and 13.6\% compared to the supervised baseline using 20\% and 50\% of the BUSI and BrEaST training data, respectively. Our research highlights that domain-inspired SSL can improve US segmentation, especially under data-limited conditions.
\end{abstract}

\begin{IEEEkeywords}
Contrastive learning, deep learning, generalisability, self-supervised learning, ultrasound imaging
\end{IEEEkeywords}

\section{Introduction}
\label{sec:introduction}
US imaging provides a non-invasive, portable and low-cost imaging solution for clinicians compared to alternative imaging modalities, such as CT and MRI. US is a popular and widely used imaging tool within clinical practice for imaging various organs, such as cardiac, breast and abdominal examination~\cite{carovac_application_2011}. However, US imaging suffers from large variability in image quality and is considered a highly operator-dependent procedure~\cite{brattain_machine_2018}. Data acquisition and interpretation of US images require significant operator skill, is time-consuming and erroneous. Using recent advancements in artificial intelligence (AI) approaches can assist clinicians in identifying key anatomical features to support clinical diagnosis, reduce human-related errors and minimise inter-operator variability. 

Deep learning (DL) approaches for US image segmentation have been extensively researched in many clinical domains. State-of-the-art supervised learning approaches have been developed using both recent transformers and convolutional neural network (CNN) based architectures~\cite{jiang_hybrid_2023, zhang_hau-net_2024, wu_cross-image_2023}. Furthermore, memory banks have been used in supervised learning to store image features~\cite{wu_cross-image_2023} or class features~\cite{yuan_unified_2024}, providing additional information to enhance image segmentation through cross-image feature aggregation~\cite{wu_cross-image_2023, yuan_unified_2024}. However, supervised learning methods are dependent on large data samples to perform effectively. 
%
These datasets are difficult to curate in the US domain with significant expertise required to acquire, anonymise, and annotate the data. Even with improvements in model development, many supervised approaches result in a steep drop in performance on unseen out-of-distribution data~\cite{liu_self-supervised_2021}. Models applied to data from alternative centres and under different imaging protocols can hinder performance~\cite{eche_toward_2021}. Furthermore, most development is focused on a single clinical domain. We recognise that US B-mode image data is commonly used across many clinical domains but with unique challenges. We hypothesise that a generalisable framework can be applied to a range of public US B-mode image datasets without a significant drop in performance on out-of-distribution data.

To overcome issues with limited annotated training data for effective supervised learning, self-supervised learning (SSL) methods are becoming increasingly popular in medical image analysis~\cite{s_azizi_big_2021, chen_self-supervised_2019}. SSL pretext learning methods learn semantically meaningful feature representations from unlabelled data. The trained model is then fine-tuned for downstream tasks, such as segmentation, using available labelled data. SSL approaches improve model performance in the downstream task with limited labelled data as well as improved generalisability to out-of-distribution datasets~\cite{z_xu_ssl-cpcd_2024, vanberlo_survey_2024}.

Several SSL approaches have been applied to US imaging, however, these are often tailored to specific clinical domain challenges, with limited generalisability assessment (e.g., wrist US~\cite{kainz_self-supervised_2023} and thyroid US~\cite{wang_thyroid_2024}). The contrastive learning approach has demonstrated improved performance for downstream tasks like segmentation~\cite{liu_self-supervised_2023} in an SSL setting. All SOTA contrastive SSL approaches rely on a transformed image to support representation learning~\cite{misra_self-supervised_2020}. Transformations often include image rotation, flipping, colour jittering or more complex augmentation approaches~\cite{leibe_unsupervised_2016}. Combining transformations has also been shown to benefit representation learning in SSL~\cite{chen_simple_2020, misra_self-supervised_2020}. However, we hypothesise that learnt representations depend on a domain-specific data-engineering technique and hence we can improve US B-mode image segmentation using domain understanding. 
Inspired by pretext-invariant representation learning (PIRL)~\cite{misra_self-supervised_2020} strategy, we propose a novel self-supervised pretext learning approach consisting of spatial and frequency-based augmentations for US B-mode images. In addition, we develop a novel loss function that aims to minimise the feature-level discrepancy and logit-level contrastive discrepancy. Key contributions of our work include:
\begin{enumerate}
    \item Domain-inspired data-engineered pretext learning with spatial and frequency-based augmentations for US B-mode images.
    \item Novel relation contrastive loss (RCL) to enhance inter-class separation. RCL compares a logit output from a shallow learnable neural network as a mean squared distance from the ground truth label. Minimising the RCL encourages the network to learn a non-linear separation between data points of negative samples while similar samples are pulled together. 
    \item Perceptual loss within contrastive SSL to weight representation learning for both higher-level and more abstract features.
    \item We provide a comprehensive benchmark of our proposed approach and baseline on 3 publicly available breast US datasets.
    \item We assess the generalisability of our approach on an unseen out-of-distribution dataset.
\end{enumerate}


\section{Related Work}
\subsection{Deep Learning for Segmentation in US B-mode Imaging}
AI techniques in US B-mode imaging are typically centred around DL model development within a supervised learning framework, often tailored to a specific clinical application. SOTA approaches for segmentation in US imaging have developed from pure convolutional approaches using CNNs. CNN-based approaches often use variants of U-Net~\cite{ronneberger_u-net_2015}. Shareef et al.~\cite{shareef_estan_2022} introduced the ESTAN network aimed at improving small breast tumour segmentation. ESTAN uses two encoder branches with different kernel shapes and sizes and three skip connections to improve multi-scale contextual information. Banerjee et al.~\cite{banerjee_ultrasound_2022} proposed SIU-Net for segmenting lumbar and thoracic bony features using their proposed inception block. This block uses multiple filter sizes with improved computational efficiency through combined $1\times1$ and $3\times3$ convolutions. They also combined features of multiple scales through dense skip connections. Meshram et al.~\cite{meshram_deep_2020} improved carotid plaque segmentation using dilated convolutional layers in a U-Net model and Qiao et al.~\cite{d_qiao_dilated_2020} used dilated convolution and squeeze excitation blocks on skip connections to improve fetal skull segmentation.

With advancements in transformer-based architectures, the latest SOTA approaches for US segmentation combine transformer and convolutional methods leveraging complementary global and local feature information, respectively. Zhang et al.~\cite{zhang_hau-net_2024} used a CNN-Transformer combination in a U-Net framework. The authors used a ResNet backbone and a novel local-global transformer block nested into skip connections to capture long-range feature information efficiently for breast US segmentation. Jiang et al.~\cite{jiang_hybrid_2023} also explored a CNN-Transformer U-Net model to improve US segmentation of the breast, thyroid and left ventricle. The authors used a coordinate residual block to extract local feature information with absolute position information and enhanced channel self-attention blocks to extract global features. Wu et al.~\cite{wu_cross-image_2023} introduced a BUSSeg model for breast US segmentation by using a parallel bi-encoder in a U-Net style architecture consisting of a transformer and CNN blocks. In addition, the authors use a cross-image dependency module to capture cross-image long-range dependencies utilising feature memory banks. 

%
These supervised learning models have demonstrated promising segmentation results across various US clinical domains. However, due to the limited availability of US data, self-supervised learning (SSL) offers a more robust approach, enabling high segmentation performance while minimizing performance degradation when applied to out-of-distribution data. Also, combined transformer and convolutional approaches~\cite{wu_cross-image_2023,jiang_hybrid_2023,zhang_hau-net_2024}  often require higher computational training and inference time that hinders clinical translation, whereas SSL techniques are independent of the model choice and can boost performance significantly.

\subsection{Self-Supervised Learning}
Self-supervised learning often follows a two-stage training strategy. Firstly, pretext learning is focused on learning representations from unlabelled data. Secondly, these learnt weights are then used in the fine-tuning downstream supervised learning tasks, such as segmentation. The objective is to learn semantically meaningful feature representations without requiring labels, thereby improving the performance of a downstream task on limited labelled datasets. Since the labels are not required during the pretext task, a large number of available unlabelled samples can be used which makes the SSL approach more generalisable to out-of-distribution samples. 

The pretext learning task is critical for developing meaningful representations of the target domain in SSL~\cite{vanberlo_survey_2024}. Often a combination of image augmentation benefits contrastive SSL pretext learning, with this unsupervised learning stage also benefiting from stronger augmentation than supervised learning~\cite{chen_simple_2020}. For example, geometric rotation transformation was applied to an image in~\cite{gidaris_unsupervised_2018} while the Jigsaw pretext task with a set of shuffled patches within an image was used in~\cite{leibe_unsupervised_2016}. The Jigsaw approach provides a strong geometric transformation to an image, a common strategy used in contrastive SSL~\cite{z_xu_ssl-cpcd_2024, xie_identification_2024, zhang_twin_2021}. The traditional Jigsaw approach learns a representation that is covariant to the perturbation, the pretext-invariant representation learning (PIRL)~\cite{misra_self-supervised_2020} approach adopts the Jigsaw task in an invariant learning strategy. 

Several SSL approaches have been established involving generative, contrastive and generative-contrastive techniques applied to medical image analysis~\cite{liu_self-supervised_2023}. The pretext learning strategy differs for each approach with a generative task focused on reconstruction, for example, recovering masked areas of an image~\cite{xie_simmim_2022}. However, the contrastive approach aims to discriminate similar and dissimilar samples~\cite {misra_self-supervised_2020}. 

Contrastive approaches are often preferred to generative approaches for downstream discriminative applications~\cite{liu_self-supervised_2023}. By avoiding low-level abstraction objectives, such as pixel-level reconstruction, contrastive learning tends to be more lightweight, as it does not require a decoder during pretext learning~\cite{liu_self-supervised_2023}. 

Widely known contrastive learning approaches include MoCo v3~\cite{he_momentum_2020}, PIRL~\cite{misra_self-supervised_2020} and BYOL~\cite{grill_bootstrap_2020}. MoCo v3 was introduced using a momentum encoder. Using a dynamic dictionary and a moving average encoder allows key feature representations to be decoupled from the minibatch size resulting in a large consistent dictionary, containing many negative samples~\cite{he_momentum_2020}. BYOL was introduced using two interacting neural networks to learn from each other. The online network predicts the target network representation, both using the same image, but under different augmented views~\cite{grill_bootstrap_2020}. PIRL introduced a
method to learn invariant representations rather than covariant representations to the pretext task used~\cite{misra_self-supervised_2020}. These approaches demonstrate improved performance in self-supervised learning. 


\subsection{Metric Learning}
Metric learning aims to learn a function to effectively compare similarities between data samples. Siamese networks~\cite{koch_siamese_2015} was used to learn a similarity function to map input pairs into a shared embedding space. The network was trained to bring similar pairs together and push dissimilar pairs apart using a linear distance metric, e.g., Euclidean distance. Prototypical networks~\cite{snell_prototypical_2017} learn an embedding that is a non-linear transformation of the input data, mapping it into an embedding space where the nearest neighbour classification is effective. The classification is based on the proximity of query instances to class prototypes in this learned embedding space. Relation Networks~\cite{f_sung_learning_2018} use an embedding module to obtain sample feature embeddings and a relation module to compute sample pair similarity. Unlike~\cite{koch_siamese_2015, snell_prototypical_2017}, the addition of the relation modules enables learning of similarity metrics in a data-driven way. 

Our work focuses on contrastive learning because it excels in discriminative downstream applications, is more lightweight during pretext learning, and favours high-level abstract feature learning compared to generative approaches \cite{liu_self-supervised_2023}. With contrastive learning dependent upon data transformations in the pretext learning task, we utilise a combination of data-specific augmentations, shown to improve SSL feature learning \cite{chen_simple_2020}. In this work, we explore novel combined spatial and frequency-based augmentation strategies aimed at US images to improve representation learning in US data. Furthermore, inspired by relation networks~\cite{f_sung_learning_2018} as a metric learning technique, we utilise relation networks and propose a novel relation contrastive loss (RCL) in a contrastive learning setting. To further guide representation learning, 
we propose to combine RCL with perceptual loss to weight feature learning with high-level features tackling high noise and poor contrast of US images. 
%
%
%
\section{Methodology}
\begin{figure*}[t!]
  \centering
  \centerline{\includegraphics[width=0.9\textwidth]{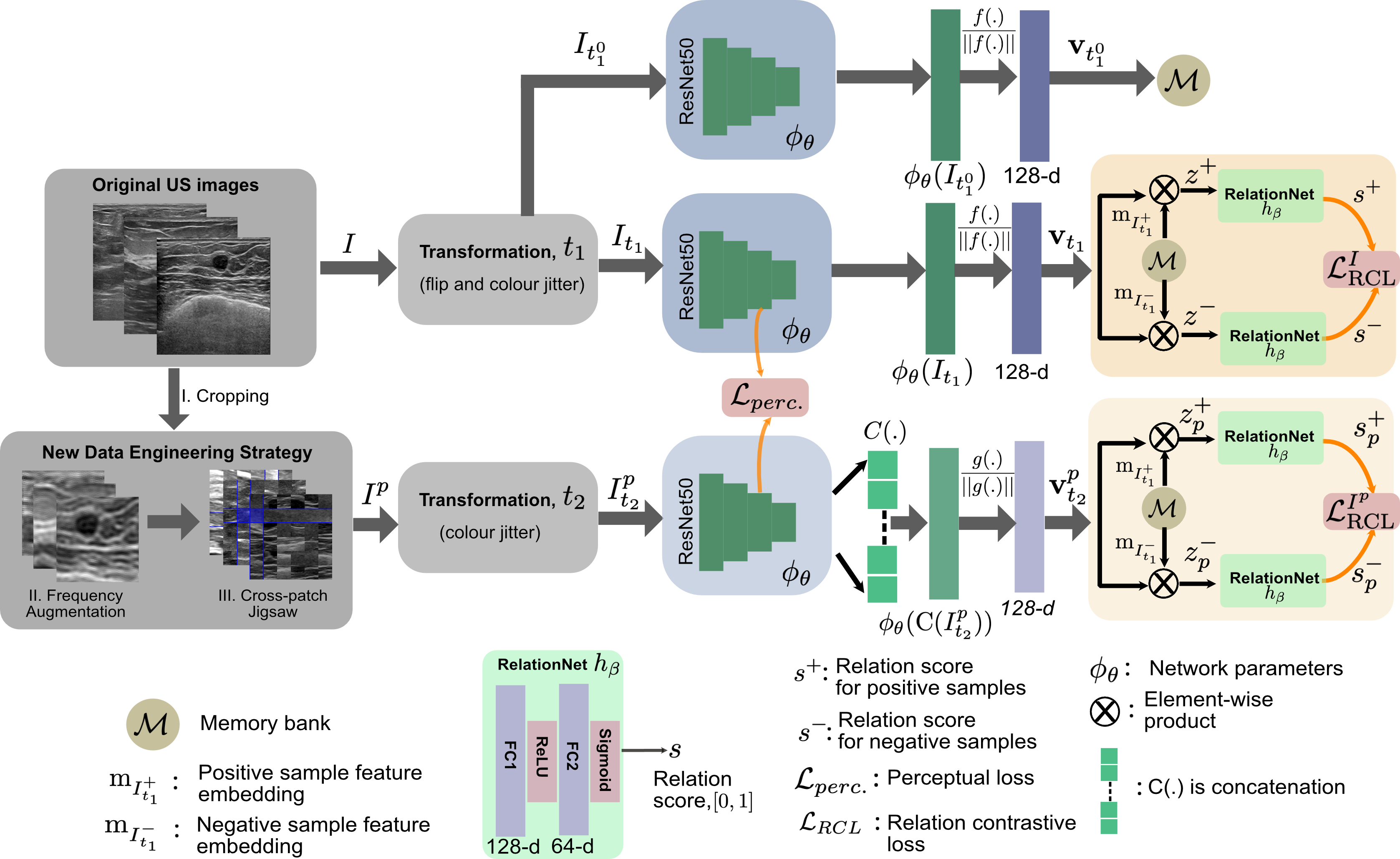}}
  \caption{Block Diagram of our proposed SSL framework with domain-inspired data engineered pretext task that integrates perceptual loss ($\mathcal{L}_{perc.}$) and novel relation contrastive loss ($\mathcal{L}_{RCL}$). A novel data engineering strategy with frequency augmentation and a proposed US data-specific Cross-patch Jigsaw is applied. 
  An ImageNet~\cite{j_deng_imagenet_2009} pre-trained ResNet50 encoder network~\cite{k_he_deep_2016} is used for our pretext task on both image-level ($I_{t_1}$) and patch-level ($I_{t^{p}_2}$). An initial representation of the images ($I^{0}_{t_1}$) from ResNet50 encoder is saved in the memory bank $\mathcal{M}$~\cite{z_wu_unsupervised_2018}. A projection network, function $f(.)$ and $g(.)$, is applied to convert the feature dimension to a $128-d$ vector. Feature embedding from patch images $I^p_{t_2}$ are concatenated. $\mathcal{L}_{RCL}$ is computed from the scores of positive, $s^{+}$ (similar) and negative, $s^{-}$ (dissimilar) samples with subscript $p$ for patch-level.}

  \label{fig:SSL_Framework}
\end{figure*}

We propose a novel self-supervised framework that explores the impact of a domain-inspired pretext task for US B-mode image data along with a novel integration of relation networks for contrastive self-supervised learning (see Fig.~\ref{fig:SSL_Framework}). We also explore the impact of perceptual loss on its ability to focus on model understanding of high-level abstract features. Our pretext task explores a novel data engineering Cross-patch Jigsaw strategy aimed at providing US B-mode image data-specific augmentation to support distinctive and meaningful representation learning during the pretext task. In addition, we apply frequency augmentations to enhance the model's ability to distinguish lesion patterns in low-contrast US images.


We propose using Relation Networks in contrastive SSL. Commonly, contrastive loss approaches, for example Noise Contrastive Estimation loss use a fixed metric like cosine similarity to determine the similarity between samples. However, we employ relation networks to compute similarity in a learnable data-driven way. This allows us to model non-linear interactions between feature embeddings enabling higher-order relationships to be computed beyond measuring just the linear alignment of feature embeddings from computing the cosine similarity. Our SSL framework complements learning meaningful feature embeddings and a robust similarity score. See Section~\ref{Relation_Network} for more details.

\subsection{Pretext Task}
\label{sec:Pretext_task}
Our pretext tasks combine a spatial and frequency component, occurring in each iteration during training. The frequency domain of a US image contains rich information on high-frequency texture variations and low-frequency tissue deformations. Our spatial transformation builds upon the Jigsaw task used in the PIRL approach~\cite{misra_self-supervised_2020}. This Jigsaw task provides a strong transformation to the image that shows high performance in self-supervised learning. We apply our frequency augmentation to a random cropped region of the image before applying our spatial patch transformation.  

\subsubsection{Frequency transformation} 
\label{sec:Frequency_transformation}
With real-world US images often suffering from a range of artefacts and noise, e.g., reverberation artefacts, speckle noise, and harmonic distortions, we utilise band stop filtering in the frequency domain to increase model robustness to noise and artefacts. The frequency domain of a US image can be obtained by computing the 2D discrete Fourier (DFT): 
%
\begin{equation}
\label{eqn:DFT_formula}
F(u, v) = \sum_{x=0}^{M-1} \sum_{y=0}^{N-1} f(x, y) \cdot e^{-j 2\pi \left( \frac{ux}{H} + \frac{vy}{W} \right)}
\end{equation}
$F(u,v)$ denotes the frequency domain value at coordinates $(u,v)$, while $f(x,y)$ denotes the spatial domain value at coordinates $(x,y)$. Here, $H$ and $W$ are the height and width of pixels of the image, respectively. According to Euler's formula $( e^{i\theta} = \cos \theta + i \sin \theta)$ Eq.~(\ref{eqn:DFT_formula}) can be written in the form:


%
\begin{align}
\label{eqn:DFT_euler}
F(u, v) = &\sum_{x=0}^{H-1} \sum_{y=0}^{W-1} U(x, y) \bigg[ \cos \left( 2\pi \left( \frac{ux}{H} + \frac{vy}{W} \right) \right) \nonumber \\
& \quad - i \sin \left( 2\pi \left( \frac{ux}{H} + \frac{vy}{W} \right) \right) \bigg]
\end{align}

\begin{figure}[H]
  \centerline{\includegraphics[width=0.95\columnwidth]{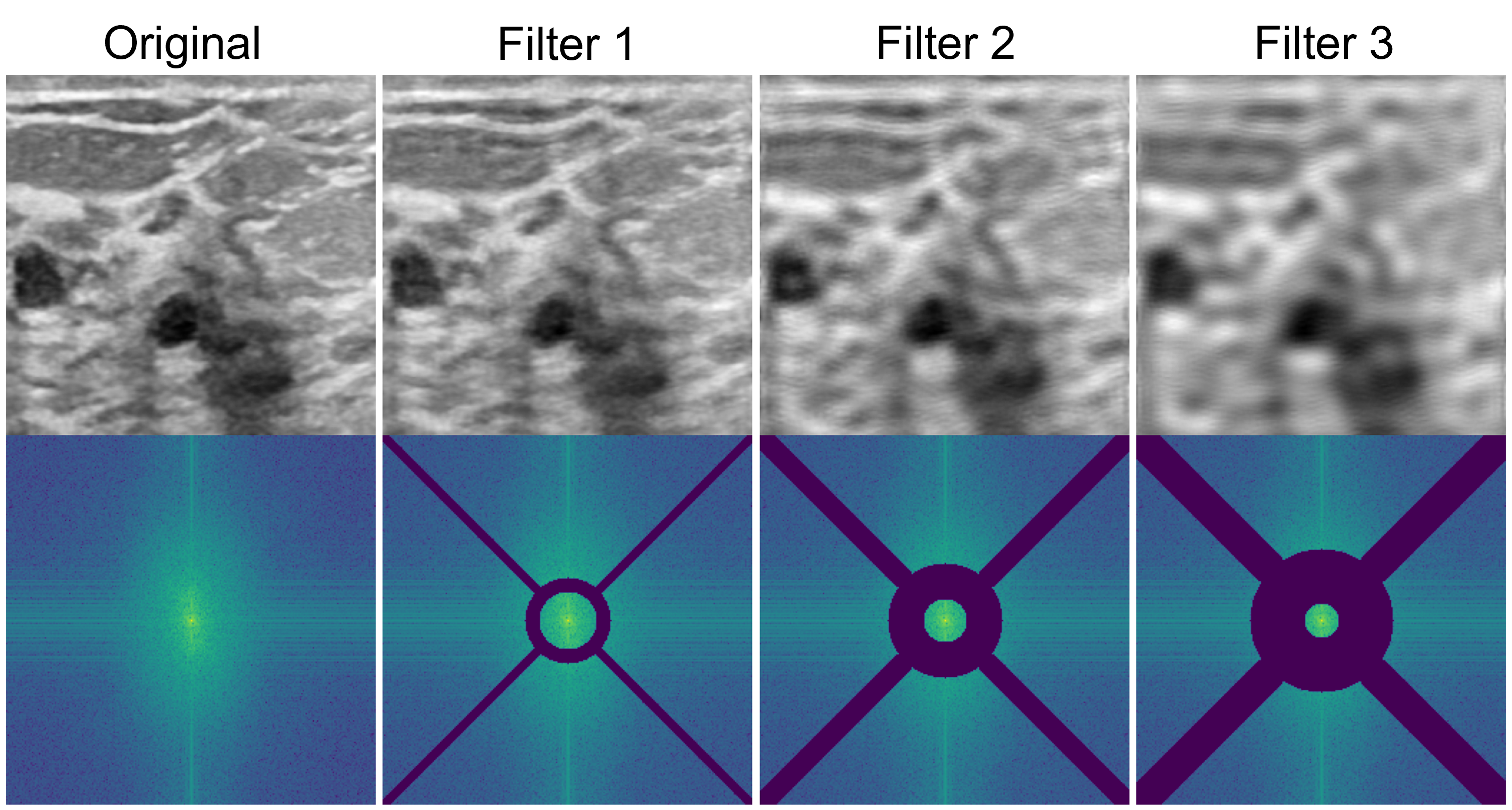}}
  \caption{Frequency-Based Filtering Augmentation: The first row shows the inverse DFT of the US image with applied filters, and the second row shows the corresponding frequency distributions. Filter settings: Original image (no filters), Filter 1 (band-stop 20-30, X-shaped filter thickness 2), Filter 2 (band-stop 15-40, X-shaped filter thickness 5), Filter 3 (band-stop 12-50, X-shaped filter thickness 8).}
  \label{fig:Freq_Augmentation}
\end{figure}

In Eq.~(\ref{eqn:DFT_euler}), real \(F_r\) (cosine term) and imaginary part \(F_i\) (sine term) can be written as \( F(u, v) = F_r(u, v) + i F_i(u, v) \). The amplitude and phase can be obtained from:
\begin{equation}
\begin{aligned}
    |F(u, v)| &= \sqrt{F_r(u, v)^2 + F_i(u, v)^2} \\
    \angle F(u, v) &= \arctan\left(\frac{F_i(u, v)}{F_r(u, v)}\right)
\end{aligned}
\end{equation}
The amplitude and phase indicate the strength and position of frequency components in the US image. We experiment with filtering frequency components within a random cropped section of the US image to distort textual information, whilst maintaining critical low-frequency components related to structures in the image. Fig. \ref{fig:Freq_Augmentation} below indicates several example augmentations applied in the frequency spectrum of a cropped US image. 

As shown these filters impact the overall image smoothness and become more blurred from left to right. This filter visually distorts structures within the image. Visually similar effects may happen during data acquisition using sub-optimal acquisition settings or poor coupling. The inner circular band stop filter has random thickness stopping frequencies within a range from radius 10 to radius 100. A minimum inner region of radius 10 is preserved for critical low-frequency information. Furthermore, we include an X-shaped band stop filter from the edge of the circular filter to apply further smoothing to the image, along the diagonal axes. We randomize the thickness of this filter from 0 to 10 and apply this filter in cases where the outer diameter of the circular filter is over 20. This threshold prevents the X-shaped filter from distorting critical visual details, ensuring meaningful image variability.

\subsubsection{Spatial transformation} 
\label{sec:spatial_transformation}
\begin{figure}[H]
  \centerline{\includegraphics[width=0.95\columnwidth]{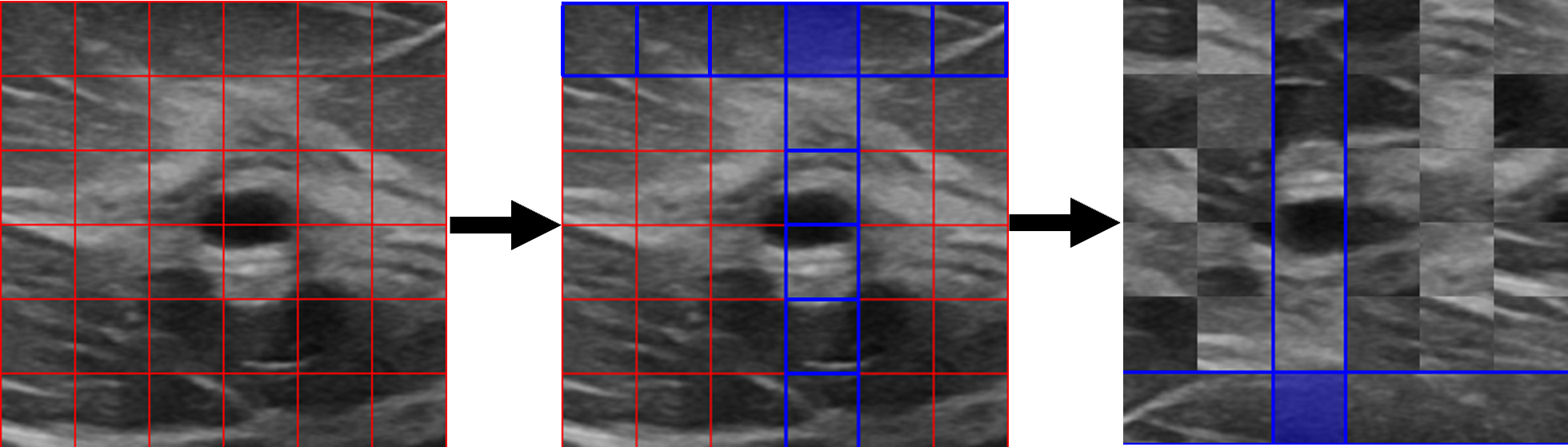}}
  \caption{Cross-patch Jigsaw Task: From left to right: Image1: Cropped frequency augmented image split into patches shown in red. Image2: Random patch selected in blue with focal patches outlined in blue and non-focal patches in red. Image3: Transformed focal and non-focal patches, focal patch area outlined in blue.}
  \label{fig:Cross_Patch_Jigsaw}
\end{figure}

A frequency-augmented random cropped area of the image $I \in \mathbb{R}^{H \times W}$ (see \ref{sec:Frequency_transformation}) is divided into 36 patches (Fig.\ref{fig:Cross_Patch_Jigsaw}). We select a random patch from these patches (shown in filled blue in Fig. \ref{fig:Cross_Patch_Jigsaw}), say $\{P_{ij}\}_{i=1, j=1}^{6,6}$, where $P_{ij}$ is the patch located at row $i$ and column $j$. 
These are then used to determine two sets of patches, namely focal patches $P_f$ and non-focal patches $P_{nf}$. $P_f$ represents patches in the same row and column as the random initial patch (outlined in blue in Fig. \ref{fig:Cross_Patch_Jigsaw}). 
\begin{equation}
P_f = \{P_{r,j} \mid j = 1, \dots, 6\} \cup \{P_{i,c} \mid i = 1, \dots, 6\}.
\end{equation}
$P_{nf}$ are the complement of focal patches (outlined in red in Fig. \ref{fig:Cross_Patch_Jigsaw}). 
\begin{equation}
P_{nf} = \{P_{ij} \mid i = 1, \dots, 6, \, j = 1, \dots, 6\} \setminus P_{nf}.
\end{equation}
All non-focal patches are shuffled ($P_{nf}^{'}$). Focal patches $P_f$ undergo horizontal and vertical flips before each patch is rotated $180^\circ$ to ensure visual coherence between focal patches. 
\begin{equation}
P_f' = \text{Rotate}_{180^\circ} \left( \text{Flip}_V \left( \text{Flip}_H(P_f) \right) \right)
\end{equation}
This spatial operation provides a strong transformation to non-focal patches while weaker augmentation to the focal patches. This operation maintains partial layer-wise structure information within the image. In Fig.~\ref{fig:SSL_Framework}, $\{P_f', P_{nf}^{'}\} \in I_{t_2}^p$.

\subsection{Feature Extraction}
Consider an ultrasound B-mode image dataset $U_D$ consisting of $N$ image samples in the training data, $U_D = \{I_1, I_2, I_3, ..., I_N\}$ and a set of image transformations, $t \in \mathcal{T}$. Original images ($I \in U_D$) undergo simple geometric and photometric transformations, referred to as $t_1$. These include random horizontal flipping, vertical flipping and colour jitter (brightness, contrast, saturation, and hue set to 0.4). $I_{t_1}$ denotes transformation $t_1$ applied to original images $I$. Patch-level transformed images ($I^p$) undergo our data-engineered strategy. The strategy includes random frequency-based filtering for a random cropped region of each image in $U_D$ and then obtaining patch-level augmented images by flipping and shuffling focal and non-focal patches (see section \ref{sec:spatial_transformation}). $I^p$ also undergoes an additional transformation ($t_2$)  that includes random colour jitter for each patch (brightness, contrast, saturation, and hue set to 0.4) denoted by $I_{t^{p}_2}$. 

A CNN with parameters $\theta$ (in our case ResNet50~\cite{k_he_deep_2016}) is used to encode image- and patch-level representations denoted as $\Phi_\theta(I_{t_1})$ and $\Phi_\theta(I_{t^{p}_2})$, respectively. At the patch level, we concatenate features from each patch, resulting in a feature vector $\phi_\theta(\text{C}(I^{p}_{t_2}))$. These image and patch-level representations are projected onto a 128-dimensional vector using a fully connected linear layer $f(.)$ and $g(.)$, respectively. These are then normalised with the $l_2$-$\text{norm}$ resulting in final feature representations at image- ($\bf{v}_{t_1}$) and patch-level ($\bf{v}_{t^{p}_2}$). 

The memory bank $\mathcal{M}$ is implemented as described in~\cite{misra_self-supervised_2020}. The $t_1$ transformed original image $I_{t^{0}_1}$ are encoded and normalised like the original images to form $\bf{v}_{t^{0}_1}$ and stored in $\mathcal{M}$. The pre-computation at each epoch enables us to utilise negative and positive sample representations in the memory bank without needing to increase the batch size to build a large sample of negative samples. These representations are updated through an exponential moving average of $\bf{v}_{t^{0}_1}$ computed in previous epochs with weight ($m_{w} = 0.5$).

\subsection{Novel Relation Contrastive Loss}
\label{Relation_Network}
We adopt a relation network (RN)~\cite{f_sung_learning_2018} with network parameters~$\beta$ (Fig.~\ref{fig:SSL_Framework}) for pretext learning. Here, we proposed to compute relation scores in a contrastive learning fashion and refer to this as ``Relation Contrastive Loss (RCL)''.

Consider a normalised positive sample representation of an image instance as $\bf{v}_{t_1}$ and a normalised target feature embedding from our moving average memory bank $\mathcal{M}$, be $\bf{v}_{t^{+}_1}$. Similarly, let $\bf{v}_{t^{-}_1}$ denote a normalised negative image sample representation from $\mathcal{M}$. We form positive ($z^+$) and negative ($z^-$) relation pairs through the element-wise product ($z^+ = \bf{v}_{t_1} \otimes \bf{v}_{t^{+}_1}$, {and} $z^- = \bf{v}_{t_1} \otimes \bf{v}_{t^{-}_1}$). RN takes each pair as input and outputs the relation scores for positive ($s^+$) and negative pairs ($s^-$), representing similarity in a range $[0-1]$ with 1 indicating a similar sample pair (see Eq.~\ref{eqn:Relation_Network}). RN consists of 2 fully connected layers with ReLU and sigmoid activation functions which take an input size of 128 and a hidden layer size of 64. 

\begin{align}
\label{eqn:Relation_Network}
s^+ &= \text{RN}(\textbf{v}_{t_1}, \textbf{v}_{t^{+}_1}) = h_\beta(z^+) \nonumber \\
s^- &= \text{RN}(\textbf{v}_{t_1}, \textbf{v}_{t^{-}_1}) = h_\beta(z^-)
\end{align}

Similarly, at patch-level (refer Fig.~\ref{fig:SSL_Framework}):
\begin{align}
\label{eqn:Relation_Network2}
s^{+}_p &= \text{RN}(\textbf{v}^p_{t_2}, \textbf{v}_{t^{+}_1}) = h_\beta(z^+_p) \nonumber \\
s^{-}_p &= \text{RN}(\textbf{v}^p_{t_2}, \textbf{v}_{t^{-}_1}) = h_\beta(z^-_p)
\end{align}

To compute relation contrastive loss (RCL, we propose to use a mean squared error (MSE) loss between relation scores $s^+$ and $s^-$ separately. Here, ground truth labels $\{1,0\}$ are compared with the relation scores. We add the computed MSE loss for positive and negative pairs with equal weighting (see Eq.~\ref{eqn:RCL}), and then average over the sample size $N$:
%
\begin{align}
\label{eqn:RCL}
\mathcal{L}_{\text{RCL}} &= \frac{1}{N} \sum_{i=1}^N \Big[ \mathcal{L}_{\text{MSE}}(s^+_i, 1) + \mathcal{L}_{\text{MSE}}(s^-_i, 0) \Big],
\end{align}
Similarly to traditional NCE, a weighted combination of RCL at the image- and patch-level is used for the final RCL loss $\mathcal{L}_{\text{RCL}}^{\text{total}}$ as below. Here, $w$ is the weight used.
\begin{align}
\label{eqn:Total_RCL}
\mathcal{L}_{\text{RCL}}^{\text{total}} &= w \cdot \mathcal{L}_{\text{RCL}}^{{I}} + (1 - w) \cdot \mathcal{L}_{\text{RCL}}^{{I^p}}
\end{align}
%


\subsection{Perceptual Loss}
%
We implement perceptual loss at layer 40 of our encoder network (ResNet50), computed between image and patch-level features. Let $\Phi_{\theta,\text{layer40}}(I_{t_1})$ denote features for the augmented original image and $\Phi_{\theta,\text{layer40}}(I{^p}_{t_2})_j$ represent features for the $j$-th patch (from 36 patches) of our patch-transformed image at network layer 40. We compute the mean squared error between image features and patch-level features as:

\begin{equation}
\mathcal{L}_{\text{perc.}} = \frac{1}{36} \sum_{j=1}^{36} \left(\Phi_{\theta,\text{layer40}}(I_{t_1}) - \Phi_{\theta,\text{layer40}}(I{^p}_{t_2})_j\right)^2
\label{eqn:percep_per_sample}
\end{equation}

The total perceptual loss is averaged across all $N$ samples:

\begin{equation}
\mathcal{L}_{\text{perc.}}^{\text{total}} = \frac{1}{N} \sum_{i=1}^N \mathcal{L}_{\text{perc.}}
\label{eqn:perceptual_loss}
\end{equation}

\subsection{Proposed Loss}
\label{losses}
Our combined loss is a weighted sum of $\mathcal{L}_{\text{RCL}}^{\text{total}}$ in Eq.~(\ref{eqn:Total_RCL}) and $\mathcal{L}_{\text{perc.}}^{\text{total}}$ in Eq.~(\ref{eqn:perceptual_loss}) with $\lambda$ denoted as weight:

\begin{align}
\label{eqn:Combined_loss}
\mathcal{L}_{\text{combined}} &= \lambda \cdot \mathcal{L}_{\text{RCL}}^{\text{total}} + (1 - \lambda) \cdot \mathcal{L}_{\text{perc.}}^{\text{total}}
\end{align}

We compare this approach with PIRL configured with perceptual loss. In this method, we also use normalised feature embeddings within a noise contrastive loss framework. The structure of the loss function used in this method is similar to the combined loss described in Eq.~(\ref{eqn:Combined_loss}), but $\mathcal{L}_{\text{RCL}}^{\text{total}}$ is replaced by $\mathcal{L}_{\text{NCE}}^{\text{total}}$. For more information on computing the total NCE loss, refer to the baseline PIRL method~\cite{misra_self-supervised_2020} and~\cite{ali_self-supervised_2023}, where normalised feature embeddings are also used. 

\section{Experiments}
\subsection{Datasets}

Three publicly available breast ultrasound datasets have been used in this study: BUSI~\cite{al-dhabyani_dataset_2020}, BrEaST~\cite{pawlowska_curated_2024} and UDIAT~\cite{yap_automated_2017}. 
BUSI was collected from $600$ women aged $25$-$75$ years old from Baheya Hospital, Egypt. BUSI contains $780$ anonymised images with $487$ benign, $210$ malignant, and $133$ normal cases. Images were acquired using a LOGIQ E9 Agile US system using a $1$-$5$ MHz ML6-15-D linear probe. Ground truth mask annotations for all images were reviewed and adjusted by expert radiologists at Baheya Hospital. 

BrEaST dataset contains $256$ anonymised breast ultrasound images from $256$ patients ($18$-$87$ years old) collected in medical centres in Poland in $2019-2022$. Benign ($154$ images), malignant ($98$ images), and normal ($4$ images) cases were collected. Four different US systems and transducers were used in the acquisition. 

UDIAT dataset B contains $163$ anonymised Breast Ultrasound images collected and labelled at the UDIAT-Centre Diagnostic, Corporacio Parc Taul, Sabadell, Spain. Benign ($110$ images) and malignant ($53$ images) breast cases were acquired using a Siemens ACUSON Sequoia C512 US system with 17L5 HD linear array probe ($8.5$ MHz).
A summary of all datasets with samples for training (train), validation (val) and test are outlined in Table \ref{tab:Experimental_settings}.

\subsection{Experimental Setup}
All methods were implemented using Pytorch and performed on a Tesla V100 GPU. A summary of the settings for the pretext learning task and the segmentation task is provided in Table \ref{tab:Experimental_settings}. We adjusted the batch size ($B$) for experiments with different datasets, due to variations in dataset sizes. To ensure experimental reproducibility, all experiments were conducted using a random seed of $42$. 
For pretext learning we trained for $2000$ epochs, using the last checkpoint for downstream task training. An SGD optimiser is used with a learning rate optimized for each method, (see Table \ref{tab:Learning_rate_effect}). All input images were resized to $224\times224$ pixels and the ResNet50 model was pre-trained on ImageNet before pretext learning.

For all downstream tasks, a polynomial $lr$ scheduler was used with initial and final learning rates of $1e^{-3}$ and $1e^{-6}$, respectively, with a learning rate decay of $0.9$. An Adam optimiser and cross-entropy loss were used for all downstream tasks. Downstream training was run for $500$ epochs with convergence occurring below $200$ epochs. We set a stopping criteria with a patience of $50$ after an initial $100$ epochs of training.

\textbf{Hyperparameters:} We investigate an optimal $\lambda$ weighting in Eq.~(\ref{eqn:Combined_loss}) for both RCL with perceptual loss and PIRL with perceptual loss. This is shown in section \ref{sec:Ablation_study}, table \ref{tab:ablation_lambda}, with $\lambda = 0.1$ and $0.75$ for these methods respectively. Our memory bank settings are the same as described in~\cite{misra_self-supervised_2020} with $m_{w}  = 0.5$ to compute the exponential moving average. Furthermore, we use $w = 0.5$ in Eq.~(\ref{eqn:Total_RCL}). This gives an equal weighting to both $\mathcal{L}_{\text{RCL}}^{\text{img}}$ and $\mathcal{L}_{\text{RCL}}^{\text{patch}}$. 


\begin{table}[]
\centering
\caption{Dataset and Experimental settings}
\label{tab:Experimental_settings}
\resizebox{\columnwidth}{!}{%
\begin{tabular}{lll}
\hline
\textbf{Dataset} & \textbf{Pretext learning} & \textbf{Downstream segmentation} \\ \hline
\begin{tabular}[c]{@{}l@{}}BUSI \\ (train: 545, \\val: 78, test: 157)\end{tabular} & \begin{tabular}[c]{@{}l@{}}$lr$: optimised as in\\ (Table \ref{tab:Learning_rate_effect}), $B=16$\\ Train: split dataset\end{tabular} & \begin{tabular}[c]{@{}l@{}}Polynomial $lr$ scheduler \\ ($0.001 \rightarrow 1e-6$), $B =16$,\\ Adam optimiser, CE loss\end{tabular} \\ \hline
\begin{tabular}[c]{@{}l@{}}UDIATdatasetB\\ (train: 113, \\val: 15, test: 35)\end{tabular} & \begin{tabular}[c]{@{}l@{}}$lr$: optimised as in \\(Table \ref{tab:Learning_rate_effect}), $B=4$\\ Train: split dataset\end{tabular} & \begin{tabular}[c]{@{}l@{}}Polynomial $lr$ scheduler \\ ($0.001 \rightarrow 1e-6$), $B=4$,\\ Adam optimiser, CE loss\end{tabular} \\ \hline
\begin{tabular}[c]{@{}l@{}}BrEaST\\ (train: 177, \\val: 24, test: 55)\end{tabular} & \begin{tabular}[c]{@{}l@{}}$lr$: optimised as in\\ (Table \ref{tab:Learning_rate_effect}), $B=8$\\ Train: split dataset\end{tabular} & \begin{tabular}[c]{@{}l@{}}Polynomial $lr$ scheduler \\ ($0.001 \rightarrow 1e-6$), $B=8$,\\ Adam optimiser, CE loss\end{tabular} \\ \hline
\begin{tabular}[c]{@{}l@{}}BUSI + BrEaST \\ (assess generalisability) \\(train: 931, val: 105, \\test: 163 (UDIAT))\end{tabular} & \begin{tabular}[c]{@{}l@{}}$lr$: optimised as in \\(Table~\ref{tab:Learning_rate_effect}), $B=32$\\ Train: BUSI + BrEaST \end{tabular} & \begin{tabular}[c]{@{}l@{}}Polynomial $lr$ scheduler \\ ($0.001 \rightarrow 1e-6$), $B=32$,\\ Adam optimiser, CE loss\end{tabular} \\ \hline
\end{tabular}%
}
\end{table}

\subsection{Evaluation Metrics}
The metrics we used to evaluate segmentation performance include: Dice score (DSC $= \frac{2 \cdot |y_{\text{pred}} \cap y_{\text{true}}|}{|y_{\text{pred}}| + |y_{\text{true}}|}$), Jaccard Coefficient (JC $= \frac{|y_{\text{pred}} \cap y_{\text{true}}|}{|y_{\text{pred}} \cup y_{\text{true}}|}$), Hausdorff distance (HD $=\max \left( \sup_{a \in y_{\text{pred}}} \inf_{b \in y_{\text{true}}} d(a, b), \sup_{b \in y_{\text{true}}} \inf_{a \in y_{\text{pred}}} d(b, a) \right)$), precision (PPV $= \frac{|y_{\text{pred}} \cap y_{\text{true}}|}{|y_{\text{pred}}|}$) and recall (Rec. $= \frac{|y_{\text{pred}} \cap y_{\text{true}}|}{|y_{\text{true}}|}$). $y_{\text{pred}}$ and $y_{\text{true}}$ represent the predicted segmentation mask and ground truth segmentation masks, respectively. The distance function $d(.)$ used is Euclidean distance for the computation of Hausdorff distance. 

\begin{table}[t!]
\centering
\caption{Learning rate hyperparameter tuning on BUSI dataset for Jigsaw pretext task. DSC is provided for each approach.}
\label{tab:Learning_rate_effect}
\begin{tabular}{l|cccccc}
\hline
\multirow{2}{*}{\textbf{Method}}                                     & \multicolumn{6}{c}{\textbf{Learning rate ($lr$)}}                                                                                                                                           \\ \cline{2-7} 
                                                                     & \multicolumn{1}{l|}{0.05}  & \multicolumn{1}{l|}{0.01}           & \multicolumn{1}{l|}{0.005} & \multicolumn{1}{l|}{0.001}          & \multicolumn{1}{l|}{0.0005}         & 0.0001         \\ \hline
\begin{tabular}[c]{@{}l@{}}PIRL\cite{misra_self-supervised_2020}\\ (baseline)\end{tabular}            & \multicolumn{1}{l|}{0.895} & \multicolumn{1}{l|}{0.869}          & \multicolumn{1}{l|}{0.849} & \multicolumn{1}{l|}{0.864}          & \multicolumn{1}{l|}{0.875}          & \textbf{0.896} \\ \hline
\begin{tabular}[c]{@{}l@{}}PIRL + \\ Perceptual \\ loss\end{tabular} & \multicolumn{1}{l|}{0.864} & \multicolumn{1}{l|}{0.879}          & \multicolumn{1}{l|}{0.870} & \multicolumn{1}{l|}{0.888}          & \multicolumn{1}{l|}{\textbf{0.867}} & 0.867          \\ \hline
RCL                                                                  & \multicolumn{1}{l|}{0.845} & \multicolumn{1}{l|}{\textbf{0.872}} & \multicolumn{1}{l|}{0.854} & \multicolumn{1}{l|}{0.868}          & \multicolumn{1}{l|}{0.856}          & 0.867          \\ \hline
\begin{tabular}[c]{@{}l@{}}RCL +\\ Perceptual \\ loss\end{tabular}   & \multicolumn{1}{l|}{0.883} & \multicolumn{1}{l|}{0.880}          & \multicolumn{1}{l|}{0.876} & \multicolumn{1}{l|}{\textbf{0.890}} & \multicolumn{1}{l|}{0.872}          & 0.873          \\ \hline
\end{tabular}
\end{table}

\section{Results}
The Res-UNet model has been used as the baseline network for supervised and self-supervised methods. We compare our novel data-engineered pretext learning strategies and loss components with fully supervised Res-UNet~\cite{Zhengxin_2018_ResUNet} and baseline Jigsaw PIRL self-supervised~\cite{misra_self-supervised_2020} approaches. The following abbreviations represent the method variations in this paper: \textbf{Jig} refers to the Jigsaw pretext task. \textbf{CP-Jig} denotes our Cross-patch Jigsaw spatial pretext task strategy (see Section \ref{sec:spatial_transformation}). \textbf{Freq} represents our frequency-based augmentation (see Section \ref{sec:Frequency_transformation}). Finally, \textbf{percep} corresponds to the perceptual loss.



\subsection{Quantitative Results on Individual datasets}
This section outlines the segmentation results of our proposed pretext learning method variations on each US dataset. 

\subsubsection{Performance on BUSI dataset}
Table \ref{tab:BUSI_optimised_results} demonstrates the segmentation performance on the held-out BUSI test set. Using the complete dataset during downstream training the top two performing methods both use Jig+Freq, using the PIRL method and RCL+percep SSL methods. These configurations obtain a dice score of $0.910$ and $0.907$, respectively, showing an improvement of $4.5\%$ and $4\%$ compared to the supervised only (Res-UNet) baseline~\cite{Zhengxin_2018_ResUNet}, but perform similarly to the PIRL (SSL) baseline~\cite{misra_self-supervised_2020} when $100\%$ of training data is used. These methods maintain the top two performances in the JC and HD. Similar gains in JC are reported over the supervised only method, whilst maintaining only slight improvement over the PIRL baseline. However, the Jig+Freq pretext task using the PIRL method shows improvements in HD by $37.6\%$ and $14.2\%$, whilst the Jig+Freq task using RCL+percep, also shows a decrease in HD by $33\%$ and $7.6\%$ compared to supervised (Res-UNet) only \cite{Zhengxin_2018_ResUNet} and PIRL \cite{misra_self-supervised_2020} baselines.

However, a greater improvement in segmentation performance can be identified using our explored data-engineered strategies and methodological changes under limited data scenarios. Using $50\%$ of training data in downstream learning, we demonstrate our frequency-based pretext learning strategy improves segmentation performance. The best performing methods are Jig+Freq pretext task in the PIRL SSL framework and CP-Jig+Freq pretext tasks using RCL respectively. Dice scores of $0.900$ and $0.901$ are achieved with half of the downstream training samples, which is an improvement of approximately $4.7\%$ compared to the supervised approach and $0.7\%$ compared to the PIRL SSL baseline. 

As training data decreases further to $20\%$ of training samples, we improve further relative to the baselines. The supervised-only (Res-UNet) \cite{Zhengxin_2018_ResUNet} method reports DSC, JC, and HD values of $0.847$, $0.834$, and $32.98$, respectively. The Jig PIRL baseline \cite{misra_self-supervised_2020} improves over the supervised-only approach with values of $0.870$, $0.856$ and $28.52$ respectively. However, RCL+percep using the Jig+Freq pretext task performs the best using $20\%$ of training samples, reporting the highest DSC, JC and HD performance of $0.882$, $0.867$ and $25.51$. This is an improvement of $1.4\%$, $1.3\%$ and $10.5\%$ compared to the SSL baseline \cite{misra_self-supervised_2020} approach.

Overall, across the BUSI dataset, we demonstrate consistently high performance in all data proportions using our frequency-based augmentation in the pretext task. We observe an improvement in the segmentation performance when combining perceptual loss either to the PIRL (PIRL+precep) or the RCL loss. 

\subsubsection{Performance on BrEaST dataset}
Segmentation results on the held-out test set of the BrEaST dataset are reported in Table \ref{tab:BrEaST_optimized_results}. Similarly to the BUSI dataset results, using $100\%$ of training samples, we report similar performance to the Jig PIRL baseline \cite{misra_self-supervised_2020}, achieving DSC, JC and HD of $0.869$, $0.849$ and $24.44$ respectively. Our best method variation and data-engineered approach is the Jig+Freq pretext task using PIRL+percep achieving similar DSC and JC. Furthermore, the SSL approaches often outperform the supervised (Res-UNet \cite{Zhengxin_2018_ResUNet}) approach (with DSC, JC and HD of $0.842$, $0.820$ and $31.60$,  respectively), demonstrating the value of pretext learning to improve downstream segmentation performance. 

As fewer samples are used during downstream training, we demonstrate the benefit of our data-engineered strategies and method developments. Using $50\%$ training samples, the top two performing methods are both RCL+percep, using the CP-Jig or CP-Jig+Freq tasks. The best-performing approach using $50\%$ training samples uses the CP-Jig pretext task, achieving segmentation results of $0.883$ (DSC), $0.864$ (JC) and $22.53$ (HD). This is an improvement over the PIRL baseline \cite{misra_self-supervised_2020} by $3\%$, $3.1\%$ and $17.6\%$, respectively. 

Using $20\%$ training samples we show that the best performance is with the RCL+percep method. However, in such a limited data scenario, the top two performing methods are jigsaw (Jig) and jigsaw combined with frequency (Jig+Freq) pretext tasks. The best-performing method is Jig+Freq with RCL+percep observing an improvement of $2.8\%$, $2.1\%$ and $18.8\%$ in DSC, JC and HD, respectively, compared to the PIRL baseline \cite{misra_self-supervised_2020}. The supervised (Res-UNet \cite{Zhengxin_2018_ResUNet}) approach shows the worst performance across $50\%$ and $20\%$ training samples compared to SSL approaches (with DSC values of 0.795 and 0.776, respectively). 

Our results on the BrEaST dataset demonstrate that methods using PIRL and PIRL with perceptual loss perform best when $100\%$ training samples are used. However, a consistent improvement is observed using RCL with perceptual loss on $50\%$ and $20\%$ limited data scenarios. Jigsaw performs well in $100\%$ and $20\%$ training scenarios while Cross-patch Jigsaw (CP-Jig) performs better at $50\%$. 

\subsubsection{Performance on UDIAT dataset} Table \ref{tab:UDIAT_optimised_results} presents the segmentation results on the held-out test set of the UDIAT dataset. The top two performing methods using all training samples are: PIRL+percep and RCL+percep, using Jig+Freq and CP-Jig+Freq tasks respectively. The best performing approach: Jig+Freq with PIRL+percep achieves DSC, JC and HD of $0.918$, $0.906$ and $16.16$, respectively. This is an improvement of $5.1\%$, $5.3\%$ and $40.5\%$ compared to the supervised (Res-UNet \cite{Zhengxin_2018_ResUNet} baseline and $1.1\%$, $1\%$ and $12.5\%$ improvement compared to the Jig PIRL \cite{misra_self-supervised_2020} baseline. 

As we reduce training samples, we maintain top performance with $50\%$ training data in both: Jig+Freq with PIRL+percep and CP-Jig+Freq with RCL+percep. The best-performing approach with $50\%$ training data is CP-Jig+Freq with RCL+percep achieving DSC, JC and HD of $0.902$, $0.890$ and $18.96$ respectively. This is an improvement of $10.7\%$, $11.3\%$ and $46.5\%$ compared to the Jig PIRL \cite{misra_self-supervised_2020} baseline. With $20\%$ training samples used in the downstream training, the top two performing methods across DSC, JC, and HD are RCL with Jig+Freq and CP-Jig. CP-Jig with RCL achieves $0.887$, $0.876$ and $21.87$ in DSC, JC and HD, respectively. This outperforms the PIRL baseline \cite{misra_self-supervised_2020} by $5.1\%$, $5.2\%$ and $33.6\%$ respectively. 

Overall, our frequency-based augmentation improves performance across all training data scenarios on the UDIAT dataset. Furthermore, perceptual loss benefits results using $100\%$ and $50\%$ training samples when combined with either PIRL or RCL. With $20\%$ training samples, the top two methods both utilise our RCL approach. 

\begin{table*}[t!h!]
\centering
\caption{Comparison of methods explored for US segmentation on BUSI dataset for different \% of training samples. All downstream models use Res-UNet with ResNet50 encoder. Here, SD is the standard deviation.}
\label{tab:BUSI_optimised_results}
\begin{tabular}{c|c|c|l|l|l|l|l}
\hline
\textbf{\begin{tabular}[c]{@{}c@{}}Pretext \\ task\end{tabular}} &
  \textbf{Method} &
  \textbf{\begin{tabular}[c]{@{}c@{}}\% train \\ samples\end{tabular}} &
  \multicolumn{1}{c|}{\textbf{DSC} $\pm$ SD} &
  \multicolumn{1}{c|}{\textbf{JC} $\pm$ SD} &
  \multicolumn{1}{c|}{\textbf{HD} $\pm$ SD} &
  \multicolumn{1}{c|}{\textbf{PPV} $\pm$ SD} &
  \multicolumn{1}{c}{\textbf{Rec.} $\pm$ SD} \\ \hline
\multirow{3}{*}{N/A} &
  \multirow{3}{*}{\begin{tabular}[c]{@{}c@{}}Res-UNet \cite{Zhengxin_2018_ResUNet}\\ (Supervised)\end{tabular}} &
  100 &
  0.871 ± 0.149 &
  0.857 ± 0.156 &
  27.40 ± 35.61 &
  0.974 ± 0.059 &
  0.877 ± 0.151 \\ \cline{3-8} 
 &
   &
  50 &
  0.860 ± 0.158 &
  0.845 ± 0.156 &
  28.21 ± 32.92 &
  0.961 ± 0.074 &
  0.877 ± 0.148 \\ \cline{3-8} 
 &
   &
  20 &
  0.847 ± 0.158 &
  0.834 ± 0.161 &
  32.98 ± 40.73 &
  0.952 ± 0.080 &
  0.874 ± 0.166 \\ \hline
\multirow{12}{*}{Jigsaw} &
  \multirow{3}{*}{\begin{tabular}[c]{@{}c@{}}PIRL \cite{misra_self-supervised_2020} \\ Baseline\end{tabular}} &
  100 &
  0.906 ± 0.123 &
  0.890 ± 0.130 &
  19.93 ± 32.05 &
  0.941 ± 0.102 &
  \textbf{0.941 ± 0.096} \\ \cline{3-8} 
 &
   &
  50 &
  0.894 ± 0.128 &
  0.879 ± 0.136 &
  21.68 ± 30.33 &
  0.960 ± 0.073 &
  0.913 ± 0.125 \\ \cline{3-8} 
 &
   &
  20 &
  0.870 ± 0.157 &
  0.856 ± 0.160 &
  28.52 ± 41.59 &
  0.946 ± 0.091 &
  0.905 ± 0.153 \\ \cline{2-8} 
 &
  \multirow{3}{*}{\begin{tabular}[c]{@{}c@{}}PIRL + \\ perceptual loss \end{tabular}} &
  100 &
  0.900 ± 0.120 &
  0.884 ± 0.130 &
  19.60 ± 26.41 &
  0.953 ± 0.073 &
  0.924 ± 0.117 \\ \cline{3-8} 
 &
   &
  50 &
  0.891 ± 0.144 &
  0.876 ± 0.150 &
  22.05 ± 35.44 &
  0.959 ± 0.064 &
  0.912 ± 0.145 \\ \cline{3-8} 
 &
   &
  20 &
  0.848 ± 0.159 &
  0.832 ± 0.163 &
  32.92 ± 40.95 &
  0.920 ± 0.138 &
  0.906 ± 0.128 \\ \cline{2-8} 
 &
  \multirow{3}{*}{\begin{tabular}[c]{@{}c@{}}RCL \end{tabular}} &
  100 &
  0.904 ± 0.125 &
  0.888 ± 0.134 &
  18.92 ± 27.92 &
  0.957 ± 0.082 &
  0.925 ± 0.114 \\ \cline{3-8} 
 &
   &
  50 &
  0.881 ± 0.146 &
  0.865 ± 0.153 &
  25.24 ± 37.31 &
  0.928 ± 0.123 &
  0.927 ± 0.110 \\ \cline{3-8} 
 &
   &
  20 &
  0.853 ± 0.166 &
  0.840 ± 0.169 &
  33.09 ± 45.47 &
  0.957 ± 0.072 &
  0.877 ± 0.175 \\ \cline{2-8} 
 &
  \multirow{3}{*}{\begin{tabular}[c]{@{}c@{}}RCL + \\ perceptual loss \end{tabular}} &
  100 &
  0.906 ± 0.125 &
  \textbf{0.891 ± 0.134} &
  18.71 ± 29.12 &
  0.962 ± 0.071 &
  0.923 ± 0.122 \\ \cline{3-8} 
 &
   &
  50 &
  0.875 ± 0.158 &
  0.862 ± 0.162 &
  27.02 ± 42.51 &
  \textbf{0.989 ± 0.031} &
  0.869 ± 0.164 \\ \cline{3-8} 
 &
   &
  20 &
  0.871 ± 0.154 &
  0.857 ± 0.160 &
  27.84 ± 39.75 &
  \textbf{0.964 ± 0.080} &
  0.888 ± 0.155 \\ \hline
\multirow{12}{*}{\begin{tabular}[c]{@{}c@{}} Jigsaw\\ + Freq\end{tabular}} &
  \multirow{3}{*}{PIRL} &
  100 &
  \textbf{0.910 ± 0.115} &
  \textbf{0.896 ± 0.124} &
  \textbf{17.09 ± 24.72} &
  0.968 ± 0.058 &
  0.921 ± 0.120 \\ \cline{3-8} 
 &
   &
  50 &
  \textbf{0.900 ± 0.133} &
  \textbf{0.886 ± 0.139} &
  \textbf{19.99 ± 30.88} &
  0.966 ± 0.066 &
  0.913 ± 0.132 \\ \cline{3-8} 
 &
   &
  20 &
  0.862 ± 0.161 &
  0.846 ± 0.166 &
  28.82 ± 40.26 &
  0.932 ± 0.123 &
  \textbf{0.911 ± 0.134} \\ \cline{2-8} 
 &
  \multirow{3}{*}{\begin{tabular}[c]{@{}c@{}}PIRL + \\ perceptual loss \end{tabular}} &
  100 &
  0.903 ± 0.121 &
  0.888 ± 0.128 &
  20.01 ± 28.93 &
  0.956 ± 0.077 &
  0.925 ± 0.112 \\ \cline{3-8} 
 &
   &
  50 &
  0.878 ± 0.146 &
  0.863 ± 0.153 &
  24.34 ± 35.13 &
  \textbf{0.980 ± 0.046} &
  0.878 ± 0.154 \\ \cline{3-8} 
 &
   &
  20 &
  0.872 ± 0.151 &
  0.857 ± 0.154 &
  27.36 ± 39.51 &
  0.953 ± 0.085 &
  0.899 ± 0.146 \\ \cline{2-8} 
 &
  \multirow{3}{*}{\begin{tabular}[c]{@{}c@{}}RCL \end{tabular}} &
  100 &
  0.897 ± 0.140 &
  0.883 ± 0.147 &
  21.11 ± 33.98 &
  \textbf{0.974 ± 0.036} &
  0.905 ± 0.151 \\ \cline{3-8} 
 &
   &
  50 &
  0.892 ± 0.136 &
  0.878 ± 0.142 &
  21.76 ± 34.15 &
  0.962 ± 0.074 &
  0.908 ± 0.132 \\ \cline{3-8} 
 &
   &
  20 &
  0.858 ± 0.170 &
  0.843 ± 0.176 &
  29.21 ± 40.06 &
  0.929 ± 0.135 &
  0.905 ± 0.130 \\ \cline{2-8} 
 &
  \multirow{3}{*}{\begin{tabular}[c]{@{}c@{}}RCL + \\ perceptual loss \end{tabular}} &
  100 &
  \textbf{0.907 ± 0.114} &
  \textbf{0.891 ± 0.124} &
  \textbf{18.42 ± 25.66} &
  0.960 ± 0.067 &
  0.924 ± 0.114 \\ \cline{3-8} 
 &
   &
  50 &
  0.885 ± 0.147 &
  0.869 ± 0.154 &
  24.53 ± 37.16 &
  0.951 ± 0.101 &
  0.910 ± 0.129 \\ \cline{3-8} 
 &
   &
  20 &
  \textbf{0.882 ± 0.140} &
  \textbf{0.867 ± 0.147} &
  \textbf{25.51 ± 34.99} &
  0.950 ± 0.078 &
  \textbf{0.911 ± 0.137} \\ \hline
\multirow{12}{*}{\begin{tabular}[c]{@{}c@{}}Cross-patch\\ Jigsaw\end{tabular}} &
  \multirow{3}{*}{PIRL} &
  100 &
  0.896 ± 0.129 &
  0.880 ± 0.136 &
  21.20 ± 30.56 &
  0.944 ± 0.092 &
  0.928 ± 0.115 \\ \cline{3-8} 
 &
   &
  50 &
  0.895 ± 0.130 &
  0.879 ± 0.141 &
  20.60 ± 29.26 &
  0.953 ± 0.078 &
  0.917 ± 0.127 \\ \cline{3-8} 
 &
   &
  20 &
  0.832 ± 0.166 &
  0.817 ± 0.172 &
  36.90 ± 42.10 &
  0.900 ± 0.128 &
  0.906 ± 0.140 \\ \cline{2-8} 
 &
  \multirow{3}{*}{\begin{tabular}[c]{@{}c@{}}PIRL + \\ perceptual loss \end{tabular}} &
  100 &
  0.899 ± 0.138 &
  0.886 ± 0.145 &
  20.67 ± 31.97 &
  \textbf{0.976 ± 0.049} &
  0.907 ± 0.142 \\ \cline{3-8} 
 &
   &
  50 &
  0.877 ± 0.138 &
  0.861 ± 0.146 &
  25.06 ± 31.78 &
  0.929 ± 0.103 &
  0.921 ± 0.125 \\ \cline{3-8} 
 &
   &
  20 &
  \textbf{0.874 ± 0.152} &
  \textbf{0.860 ± 0.158} &
  27.50 ± 40.82 &
  \textbf{0.961 ± 0.072} &
  0.893 ± 0.156 \\ \cline{2-8} 
 &
  \multirow{3}{*}{\begin{tabular}[c]{@{}c@{}}RCL\end{tabular}} &
  100 &
  0.893 ± 0.134 &
  0.879 ± 0.141 &
  22.82 ± 34.64 &
  0.965 ± 0.085 &
  0.905 ± 0.127 \\ \cline{3-8} 
 &
   &
  50 &
  0.880 ± 0.138 &
  0.863 ± 0.146 &
  25.75 ± 35.40 &
  0.955 ± 0.098 &
  0.897 ± 0.123 \\ \cline{3-8} 
 &
   &
  20 &
  0.852 ± 0.160 &
  0.838 ± 0.164 &
  31.12 ± 39.47 &
  0.951 ± 0.081 &
  0.878 ± 0.163 \\ \cline{2-8} 
 &
  \multirow{3}{*}{\begin{tabular}[c]{@{}c@{}}RCL + \\ perceptual loss \end{tabular}} &
  100 &
  0.896 ± 0.132 &
  0.881 ± 0.138 &
  20.85 ± 33.35 &
  0.951 ± 0.068 &
  0.924 ± 0.136 \\ \cline{3-8} 
 &
   &
  50 &
  0.881 ± 0.133 &
  0.863 ± 0.142 &
  24.95 ± 32.44 &
  0.926 ± 0.108 &
  \textbf{0.931 ± 0.111} \\ \cline{3-8} 
 &
   &
  20 &
  0.872 ± 0.155 &
  0.859 ± 0.158 &
  \textbf{26.83 ± 39.88} &
  0.954 ± 0.078 &
  0.899 ± 0.155 \\ \hline
\multirow{12}{*}{\begin{tabular}[c]{@{}c@{}}Cross-patch\\ Jigsaw \\+ Freq\end{tabular}} &
  \multirow{3}{*}{PIRL} &
  100 &
  0.897 ± 0.129 &
  0.882 ± 0.137 &
  20.56 ± 27.83 &
  0.960 ± 0.076 &
  0.917 ± 0.123 \\ \cline{3-8} 
 &
   &
  50 &
  0.884 ± 0.149 &
  0.869 ± 0.157 &
  22.70 ± 31.90 &
  0.960 ± 0.088 &
  0.903 ± 0.143 \\ \cline{3-8} 
 &
   &
  20 &
  0.864 ± 0.149 &
  0.849 ± 0.155 &
  28.17 ± 36.95 &
  0.942 ± 0.084 &
  0.897 ± 0.149 \\ \cline{2-8} 
 &
  \multirow{3}{*}{\begin{tabular}[c]{@{}c@{}}PIRL + \\ perceptual loss \end{tabular}} &
  100 &
  0.890 ± 0.141 &
  0.874 ± 0.149 &
  23.71 ± 36.84 &
  0.936 ± 0.121 &
  \textbf{0.931 ± 0.104} \\ \cline{3-8} 
 &
   &
  50 &
  0.898 ± 0.140 &
  \textbf{0.884 ± 0.147} &
  21.20 ± 35.92 &
  0.968 ± 0.067 &
  0.908 ± 0.141 \\ \cline{3-8} 
 &
   &
  20 &
  0.858 ± 0.148 &
  0.844 ± 0.153 &
  30.21 ± 36.79 &
  0.938 ± 0.103 &
  0.897 ± 0.141 \\ \cline{2-8} 
 &
  \multirow{3}{*}{\begin{tabular}[c]{@{}c@{}}RCL \end{tabular}} &
  100 &
  0.905 ± 0.122 &
  0.890 ± 0.131 &
  18.58 ± 27.73 &
  0.962 ± 0.075 &
  0.923 ± 0.114 \\ \cline{3-8} 
 &
   &
  50 &
  \textbf{0.901 ± 0.120} &
  \textbf{0.884 ± 0.129} &
  \textbf{19.55 ± 28.08} &
  0.942 ± 0.091 &
  \textbf{0.933 ± 0.104} \\ \cline{3-8} 
 &
   &
  20 &
  0.865 ± 0.156 &
  0.851 ± 0.160 &
  29.07 ± 39.36 &
  0.938 ± 0.097 &
  0.906 ± 0.147 \\ \cline{2-8} 
 &
  \multirow{3}{*}{\begin{tabular}[c]{@{}c@{}}RCL + \\ perceptual loss \end{tabular}} &
  100 &
  0.896 ± 0.134 &
  0.881 ± 0.141 &
  21.55 ± 32.79 &
  0.962 ± 0.061 &
  0.914 ± 0.137 \\ \cline{3-8} 
 &
   &
  50 &
  0.883 ± 0.142 &
  0.868 ± 0.150 &
  24.21 ± 35.08 &
  0.953 ± 0.094 &
  0.908 ± 0.132 \\ \cline{3-8} 
 &
   &
  20 &
  0.837 ± 0.168 &
  0.823 ± 0.171 &
  36.38 ± 46.18 &
  0.920 ± 0.111 &
  0.897 ± 0.170 \\ \hline
  
\end{tabular}
\end{table*}

\begin{table*}[t!h!]
\centering
\caption{Comparison of methods explored for US segmentation on BrEaST dataset for different \% of training samples. All downstream models use Res-UNet with ResNet50 encoder. Here, SD is the standard deviation.}
\label{tab:BrEaST_optimized_results}
\begin{tabular}{c|c|c|l|l|l|l|l}
\hline
\textbf{\begin{tabular}[c]{@{}c@{}}Pretext \\ task\end{tabular}} &
  \textbf{Method} &
  \textbf{\begin{tabular}[c]{@{}c@{}}\% train \\ samples\end{tabular}} &
  \multicolumn{1}{c|}{\textbf{DSC} $\pm$ SD} &
  \multicolumn{1}{c|}{\textbf{JC} $\pm$ SD} &
  \multicolumn{1}{c|}{\textbf{HD} $\pm$ SD} &
  \multicolumn{1}{c|}{\textbf{PPV} $\pm$ SD} &
  \multicolumn{1}{c}{\textbf{Rec.} $\pm$ SD} \\ \hline
 &
   &
  100 &
  0.842 ± 0.132 &
  0.820 ± 0.137 &
  31.60 ± 31.12 &
  0.910 ± 0.115 &
  0.899 ± 0.088 \\ \cline{3-8} 
 &
   &
  50 &
  0.795 ± 0.120 &
  0.773 ± 0.124 &
  40.87 ± 27.93 &
  0.907 ± 0.109 &
  0.853 ± 0.110 \\ \cline{3-8} 
\multirow{-3}{*}{N/A} &
  \multirow{-3}{*}{\begin{tabular}[c]{@{}c@{}}Res-UNet \cite{Zhengxin_2018_ResUNet}\\ (Supervised)\end{tabular}} &
  20 &
  0.776 ± 0.147 &
  0.765 ± 0.146 &
  50.88 ± 39.22 &
  0.974 ± 0.045 &
  0.785 ± 0.155 \\ \hline
 &
   &
  100 &
  \textbf{0.869 ± 0.115} &
  \textbf{0.849 ± 0.120} &
  \textbf{24.44 ± 24.15} &
  0.954 ± 0.063 &
  0.887 ± 0.113 \\ \cline{3-8} 
 &
   &
  50 &
  0.858 ± 0.131 &
  0.838 ± 0.133 &
  27.33 ± 29.48 &
  0.929 ± 0.100 &
  0.901 ± 0.095 \\ \cline{3-8} 
 &
  \multirow{-3}{*}{\begin{tabular}[c]{@{}c@{}}PIRL \cite{misra_self-supervised_2020} \\ Baseline\end{tabular}} &
  20 &
  0.811 ± 0.160 &
  0.799 ± 0.157 &
  40.84 ± 40.68 &
  0.971 ± 0.045 &
  0.820 ± 0.161 \\ \cline{2-8} 
 &
   &
  100 &
  0.860 ± 0.104 &
  0.839 ± 0.110 &
  \textbf{24.67 ± 21.15} &
  0.936 ± 0.069 &
  0.891 ± 0.109 \\ \cline{3-8} 
 &
   &
  50 &
  0.867 ± 0.115 &
  \textbf{0.849 ± 0.120} &
  26.54 ± 28.37 &
  \textbf{0.977 ± 0.038} &
  0.865 ± 0.119 \\ \cline{3-8} 
 &
  \multirow{-3}{*}{\begin{tabular}[c]{@{}c@{}}PIRL + \\ perceptual loss \end{tabular}} &
  20 &
  0.783 ± 0.155 &
  0.768 ± 0.153 &
  46.20 ± 39.31 &
  0.942 ± 0.079 &
  0.818 ± 0.162 \\ \cline{2-8} 
 &
   &
  100 &
  0.802 ± 0.139 &
  0.782 ± 0.141 &
  39.84 ± 33.59 &
  0.942 ± 0.077 &
  0.833 ± 0.142 \\ \cline{3-8} 
 &
   &
  50 &
  0.840 ± 0.123 &
  0.821 ± 0.125 &
  32.20 ± 29.37 &
  0.919 ± 0.093 &
  0.894 ± 0.101 \\ \cline{3-8} 
 &
  \multirow{-3}{*}{\begin{tabular}[c]{@{}c@{}}RCL \end{tabular}} &
  20 &
  0.759 ± 0.142 &
  0.751 ± 0.142 &
  56.49 ± 38.69 &
  0.985 ± 0.035 &
  0.762 ± 0.147 \\ \cline{2-8} 
 &
   &
  100 &
  0.864 ± 0.114 &
  0.845 ± 0.116 &
  26.50 ± 26.74 &
  \textbf{0.962 ± 0.045} &
  0.874 ± 0.118 \\ \cline{3-8} 
 &
   &
  50 &
  0.855 ± 0.119 &
  0.835 ± 0.124 &
  28.41 ± 28.20 &
  0.949 ± 0.078 &
  0.878 ± 0.113 \\ \cline{3-8} 
\multirow{-12}{*}{Jigsaw} &
  \multirow{-3}{*}{\begin{tabular}[c]{@{}c@{}}RCL + \\ perceptual loss \end{tabular}} &
  20 &
  \textbf{0.833 ± 0.139} &
  \textbf{0.816 ± 0.141} &
  34.06 ± 33.23 &
  0.963 ± 0.047 &
  0.846 ± 0.147 \\ \hline
 &
   &
  100 &
  0.847 ± 0.130 &
  0.828 ± 0.133 &
  29.33 ± 27.33 &
  0.951 ± 0.084 &
  0.869 ± 0.121 \\ \cline{3-8} 
 &
   &
  50 &
  0.844 ± 0.158 &
  0.825 ± 0.160 &
  30.88 ± 37.08 &
  0.913 ± 0.130 &
  0.904 ± 0.103 \\ \cline{3-8} 
 &
  \multirow{-3}{*}{PIRL} &
  20 &
  0.819 ± 0.143 &
  0.799 ± 0.143 &
  36.83 ± 35.60 &
  0.919 ± 0.094 &
  0.869 ± 0.144 \\ \cline{2-8} 
 &
   &
  100 &
  \textbf{0.868 ± 0.114} &
  \textbf{0.850 ± 0.115} &
  25.75 ± 25.55 &
  0.941 ± 0.081 &
  0.896 ± 0.105 \\ \cline{3-8} 
 &
   &
  50 &
  0.868 ± 0.108 &
  0.848 ± 0.112 &
  25.91 ± 27.02 &
  0.938 ± 0.079 &
  0.901 ± 0.103 \\ \cline{3-8} 
 &
  \multirow{-3}{*}{\begin{tabular}[c]{@{}c@{}}PIRL + \\ perceptual loss \end{tabular}} &
  20 &
  0.823 ± 0.156 &
  0.808 ± 0.154 &
  37.38 ± 39.71 &
  0.973 ± 0.032 &
  0.826 ± 0.157 \\ \cline{2-8} 
 &
   &
  100 &
  0.859 ± 0.109 &
  0.839 ± 0.113 &
  27.36 ± 24.64 &
  0.958 ± 0.044 &
  0.874 ± 0.119 \\ \cline{3-8} 
 &
   &
  50 &
  0.786 ± 0.167 &
  0.776 ± 0.165 &
  48.92 ± 44.25 &
  \textbf{0.974 ± 0.052} &
  0.793 ± 0.166 \\ \cline{3-8} 
 &
  \multirow{-3}{*}{\begin{tabular}[c]{@{}c@{}}RCL \end{tabular}} &
  20 &
  0.828 ± 0.134 &
  0.808 ± 0.136 &
  \textbf{31.84 ± 28.26} &
  0.938 ± 0.068 &
  0.859 ± 0.135 \\ \cline{2-8} 
 &
   &
  100 &
  0.862 ± 0.114 &
  0.844 ± 0.117 &
  26.52 ± 24.54 &
  0.953 ± 0.063 &
  0.881 ± 0.124 \\ \cline{3-8} 
 &
   &
  50 &
  0.834 ± 0.137 &
  0.815 ± 0.141 &
  33.11 ± 32.45 &
  0.934 ± 0.084 &
  0.868 ± 0.136 \\ \cline{3-8} 
\multirow{-12}{*}{\begin{tabular}[c]{@{}c@{}} Jigsaw\\ + Freq\end{tabular}} &
  \multirow{-3}{*}{\begin{tabular}[c]{@{}c@{}}RCL + \\ perceptual loss \end{tabular}} &
  20 &
  \textbf{0.834 ± 0.137} &
  \textbf{0.816 ± 0.141} &
  \textbf{33.18 ± 31.78} &
  0.953 ± 0.072 &
  0.857 ± 0.140 \\ \hline
 &
   &
  100 &
  0.860 ± 0.108 &
  0.840 ± 0.113 &
  26.68 ± 23.59 &
  0.952 ± 0.062 &
  0.878 ± 0.108 \\ \cline{3-8} 
 &
   &
  50 &
  0.857 ± 0.153 &
  0.838 ± 0.155 &
  28.08 ± 34.58 &
  0.944 ± 0.103 &
  0.886 ± 0.118 \\ \cline{3-8} 
 &
  \multirow{-3}{*}{PIRL} &
  20 &
  0.819 ± 0.145 &
  0.799 ± 0.149 &
  35.35 ± 31.40 &
  0.904 ± 0.114 &
  \textbf{0.878 ± 0.128} \\ \cline{2-8} 
 &
   &
  100 &
  0.865 ± 0.103 &
  0.845 ± 0.108 &
  26.21 ± 24.92 &
  0.956 ± 0.058 &
  0.880 ± 0.096 \\ \cline{3-8} 
 &
   &
  50 &
  0.835 ± 0.147 &
  0.815 ± 0.148 &
  34.07 ± 35.70 &
  0.946 ± 0.104 &
  0.863 ± 0.113 \\ \cline{3-8} 
 &
  \multirow{-3}{*}{\begin{tabular}[c]{@{}c@{}}PIRL + \\ perceptual loss \end{tabular}} &
  20 &
  0.830 ± 0.154 &
  \textbf{0.816 ± 0.154} &
  36.39 ± 37.58 &
  0.975 ± 0.045 &
  0.835 ± 0.156 \\ \cline{2-8} 
 &
   &
  100 &
  0.861 ± 0.114 &
  0.839 ± 0.120 &
  26.95 ± 25.87 &
  0.921 ± 0.103 &
  \textbf{0.905 ± 0.099} \\ \cline{3-8} 
 &
   &
  50 &
  0.850 ± 0.117 &
  0.829 ± 0.112 &
  28.38 ± 25.35 &
  0.913 ± 0.082 &
  0.902 ± 0.102 \\ \cline{3-8} 
 &
  \multirow{-3}{*}{\begin{tabular}[c]{@{}c@{}}RCL \end{tabular}} &
  20 &
  0.788 ± 0.150 &
  0.770 ± 0.151 &
  43.33 ± 35.07 &
  0.929 ± 0.102 &
  0.826 ± 0.144 \\ \cline{2-8} 
 &
   &
  100 &
  0.862 ± 0.112 &
  0.842 ± 0.114 &
  25.71 ± 23.31 &
  0.921 ± 0.084 &
  \textbf{0.906 ± 0.104} \\ \cline{3-8} 
 &
   &
  50 &
  \textbf{0.883 ± 0.103} &
  \textbf{0.864 ± 0.107} &
  \textbf{22.53 ± 24.67} &
  0.950 ± 0.059 &
  \textbf{0.906 ± 0.095} \\ \cline{3-8} 
\multirow{-12}{*}{\begin{tabular}[c]{@{}c@{}}Cross-patch\\ Jigsaw\end{tabular}} &
  \multirow{-3}{*}{\begin{tabular}[c]{@{}c@{}}RCL + \\ perceptual loss \end{tabular}} &
  20 &
  0.818 ± 0.132 &
  0.799 ± 0.134 &
  35.04 ± 30.42 &
  0.928 ± 0.087 &
  0.860 ± 0.139 \\ \hline
 &
   &
  100 &
  0.867 ± 0.107 &
  0.847 ± 0.114 &
  24.77 ± 22.65 &
  0.951 ± 0.066 &
  0.886 ± 0.104 \\ \cline{3-8} 
 &
   &
  50 &
  0.818 ± 0.142 &
  0.800 ± 0.143 &
  38.06 ± 35.19 &
  0.963 ± 0.063 &
  0.832 ± 0.141 \\ \cline{3-8} 
 &
  \multirow{-3}{*}{PIRL} &
  20 &
  0.819 ± 0.131 &
  0.798 ± 0.136 &
  33.78 ± 26.80 &
  0.915 ± 0.103 &
  \textbf{0.870 ± 0.119} \\ \cline{2-8} 
 &
   &
  100 &
  0.855 ± 0.118 &
  0.838 ± 0.120 &
  28.99 ± 27.36 &
  \textbf{0.974 ± 0.048} &
  0.860 ± 0.122 \\ \cline{3-8} 
 &
   &
  50 &
  0.820 ± 0.171 &
  0.801 ± 0.171 &
  39.04 ± 46.72 &
  0.893 ± 0.168 &
  0.899 ± 0.089 \\ \cline{3-8} 
 &
  \multirow{-3}{*}{\begin{tabular}[c]{@{}c@{}}PIRL + \\ perceptual loss \end{tabular}} &
  20 &
  0.811 ± 0.162 &
  0.797 ± 0.162 &
  41.24 ± 40.98 &
  \textbf{0.991 ± 0.025} &
  0.804 ± 0.161 \\ \cline{2-8} 
 &
   &
  100 &
  0.816 ± 0.136 &
  0.795 ± 0.138 &
  37.87 ± 30.70 &
  0.886 ± 0.129 &
  0.898 ± 0.098 \\ \cline{3-8} 
 &
   &
  50 &
  0.841 ± 0.144 &
  0.819 ± 0.148 &
  32.96 ± 37.09 &
  0.904 ± 0.130 &
  \textbf{0.906 ± 0.093} \\ \cline{3-8} 
 &
  \multirow{-3}{*}{\begin{tabular}[c]{@{}c@{}}RCL \end{tabular}} &
  20 &
   0.795 ± 0.161 &
   0.785 ± 0.158 &
   46.94 ± 43.03 &
  \textbf{0.990 ± 0.020} &
  0.791 ± 0.163 \\ \cline{2-8} 
 &
   &
  100 &
  0.841 ± 0.119 &
  0.821 ± 0.120 &
  30.66 ± 25.72 &
  0.916 ± 0.071 &
  0.893 ± 0.115 \\ \cline{3-8} 
 &
   &
  50 &
  \textbf{0.870 ± 0.112} &
  \textbf{0.849 ± 0.118} &
  \textbf{25.11 ± 24.75} &
  0.953 ± 0.069 &
  0.890 ± 0.095 \\ \cline{3-8} 
\multirow{-12}{*}{\begin{tabular}[c]{@{}c@{}}Cross-patch\\ Jigsaw \\+ Freq\end{tabular}} &
  \multirow{-3}{*}{\begin{tabular}[c]{@{}c@{}} RCL +\\ perceptual loss \end{tabular}} &
  20 &
  0.820 ± 0.146 &
  0.804 ± 0.145 &
  37.41 ± 36.61 &
  0.968 ± 0.050 &
  0.828 ± 0.146 \\ \hline
  
\end{tabular}
\end{table*}

\begin{table*}[t!h!]
\centering
\caption{Comparison of methods explored for US segmentation on UDIAT dataset for different \% of training samples. All downstream models use Res-UNet with ResNet50 encoder. Here, SD is the standard deviation.}
\label{tab:UDIAT_optimised_results}
\begin{tabular}{c|c|c|l|l|l|l|l}
\hline
\textbf{\begin{tabular}[c]{@{}c@{}}Pretext \\ task\end{tabular}} &
  \textbf{Method} &
  \textbf{\begin{tabular}[c]{@{}c@{}}\% train \\ samples\end{tabular}} &
  \multicolumn{1}{c|}{\textbf{DSC} $\pm$ SD} &
  \multicolumn{1}{c|}{\textbf{JC} $\pm$ SD} &
  \multicolumn{1}{c|}{\textbf{HD} $\pm$ SD} &
  \multicolumn{1}{c|}{\textbf{PPV} $\pm$ SD} &
  \multicolumn{1}{c}{\textbf{Rec.} $\pm$ SD} \\ \hline
 &
   &
  100 &
  0.873 ± 0.116 &
  0.860 ± 0.123 &
  27.18 ± 30.13 &
  0.919 ± 0.098 &
  0.932 ± 0.102 \\ \cline{3-8} 
 &
   &
  50 &
  0.865 ± 0.136 &
  0.852 ± 0.140 &
  26.60 ± 31.88 &
  0.935 ± 0.101 &
  0.912 ± 0.102 \\ \cline{3-8}
\multirow{-3}{*}{N/A} &
  \multirow{-3}{*}{\begin{tabular}[c]{@{}c@{}}Res-UNet \cite{Zhengxin_2018_ResUNet}\\ (Supervised)\end{tabular}} &
  20 &
  0.823 ± 0.137 &
  0.808 ± 0.143 &
  38.15 ± 36.30 &
  0.874 ± 0.131 &
  0.924 ± 0.093 \\ \hline
 &
   &
  100 &
  0.908 ± 0.108 &
  0.897 ± 0.113 &
  18.47 ± 26.59 &
  0.956 ± 0.065 &
  0.937 ± 0.103 \\ \cline{3-8} 
 &
   &
  50 &
  0.815 ± 0.153 &
  0.800 ± 0.159 &
  35.46 ± 33.13 &
  0.859 ± 0.142 &
  0.933 ± 0.090 \\ \cline{3-8} 
 &
  \multirow{-3}{*}{\begin{tabular}[c]{@{}c@{}}PIRL \cite{misra_self-supervised_2020} \\ Baseline\end{tabular}} &
  20 &
  0.844 ± 0.116 &
  0.833 ± 0.122 &
  32.92 ± 30.45 &
  0.953 ± 0.080 &
  0.877 ± 0.114 \\ \cline{2-8} 
 &
   &
  100 &
  0.905 ± 0.113 &
  0.893 ± 0.119 &
  18.05 ± 27.66 &
  0.948 ± 0.081 &
  0.940 ± 0.100 \\ \cline{3-8} 
 &
   &
  50 &
  0.849 ± 0.150 &
  0.834 ± 0.155 &
  32.68 ± 39.11 &
  0.894 ± 0.122 &
  \textbf{0.935 ± 0.094} \\ \cline{3-8} 
 &
  \multirow{-3}{*}{\begin{tabular}[c]{@{}c@{}}PIRL + \\ perceptual loss \end{tabular}} &
  20 &
  \textbf{0.884 ± 0.106} &
  \textbf{0.871 ± 0.111} &
  23.44 ± 26.92 &
  0.941 ± 0.069 &
  0.927 ± 0.094 \\ \cline{2-8} 
 &
   &
  100 &
  0.890 ± 0.127 &
  0.879 ± 0.132 &
  22.78 ± 32.05 &
  0.956 ± 0.082 &
  0.921 ± 0.116 \\ \cline{3-8} 
 &
   &
  50 &
  0.896 ± 0.108 &
  0.885 ± 0.115 &
  20.09 ± 28.63 &
  0.964 ± 0.050 &
  0.915 ± 0.119 \\ \cline{3-8} 
 &
  \multirow{-3}{*}{\begin{tabular}[c]{@{}c@{}}RCL \end{tabular}} &
  20 &
  0.855 ± 0.105 &
  0.842 ± 0.111 &
  28.19 ± 26.22 &
  0.913 ± 0.089 &
  0.926 ± 0.096 \\ \cline{2-8} 
 &
   &
  100 &
  0.906 ± 0.107 &
  0.895 ± 0.111 &
  18.78 ± 27.45 &
  \textbf{0.978 ± 0.032} &
  0.914 ± 0.117 \\ \cline{3-8} 
 &
   &
  50 &
  0.890 ± 0.107 &
  0.877 ± 0.112 &
  21.94 ± 27.32 &
  0.958 ± 0.053 &
  0.915 ± 0.106 \\ \cline{3-8} 
\multirow{-12}{*}{Jigsaw} &
  \multirow{-3}{*}{\begin{tabular}[c]{@{}c@{}}RCL + \\ perceptual loss \end{tabular}} &
  20 &
  0.870 ± 0.114 &
  0.856 ± 0.119 &
  26.42 ± 29.09 &
  0.938 ± 0.080 &
  0.915 ± 0.097 \\ \hline
 &
   &
  100 &
  0.900 ± 0.110 &
  0.887 ± 0.116 &
  19.86 ± 27.59 &
  0.929 ± 0.081 &
  \textbf{0.954 ± 0.083} \\ \cline{3-8} 
 &
   &
  50 &
  0.882 ± 0.107 &
  0.868 ± 0.112 &
  23.09 ± 25.87 &
  0.949 ± 0.068 &
  0.915 ± 0.101 \\ \cline{3-8} 
 &
  \multirow{-3}{*}{PIRL} &
  20 &
  0.869 ± 0.116 &
  0.856 ± 0.123 &
  25.63 ± 27.96 &
  0.925 ± 0.097 &
  0.927 ± 0.101 \\ \cline{2-8} 
 &
   &
  100 &
  \textbf{0.918 ± 0.097} &
  \textbf{0.906 ± 0.102} &
  \textbf{16.16 ± 24.81} &
  0.966 ± 0.049 &
  0.938 ± 0.103 \\ \cline{3-8} 
 &
   &
  50 &
  \textbf{0.900 ± 0.109} &
  \textbf{0.887 ± 0.114} &
  19.42 ± 26.86 &
  0.962 ± 0.053 &
  0.920 ± 0.101 \\ \cline{3-8} 
 &
  \multirow{-3}{*}{\begin{tabular}[c]{@{}c@{}}PIRL + \\ perceptual loss \end{tabular}} &
  20 &
  0.858 ± 0.119 &
  0.844 ± 0.125 &
  28.60 ± 30.64 &
  0.917 ± 0.086 &
  0.920 ± 0.103 \\ \cline{2-8} 
 &
   &
  100 &
  0.901 ± 0.103 &
  0.890 ± 0.109 &
  19.89 ± 26.36 &
  0.949 ± 0.073 &
  0.939 ± 0.103 \\ \cline{3-8} 
 &
   &
  50 &
  0.709 ± 0.209 &
  0.691 ± 0.214 &
  71.36 ± 62.44 &
  0.743 ± 0.212 &
  \textbf{0.942 ± 0.085} \\ \cline{3-8} 
 &
  \multirow{-3}{*}{\begin{tabular}[c]{@{}c@{}} RCL \end{tabular}} &
  20 &
  \textbf{0.884 ± 0.114} &
  \textbf{0.871 ± 0.121} &
  \textbf{22.53 ± 27.64} &
  0.940 ± 0.072 &
  0.923 ± 0.100 \\ \cline{2-8} 
 &
   &
  100 &
  0.889 ± 0.121 &
  0.875 ± 0.126 &
  20.98 ± 28.41 &
  0.935 ± 0.093 &
  0.935 ± 0.091 \\ \cline{3-8} 
 &
   &
  50 &
  0.889 ± 0.107 &
  0.875 ± 0.114 &
  22.75 ± 26.97 &
  0.959 ± 0.063 &
  0.912 ± 0.106 \\ \cline{3-8} 
\multirow{-12}{*}{\begin{tabular}[c]{@{}c@{}} Jigsaw\\ + Freq\end{tabular}} &
  \multirow{-3}{*}{\begin{tabular}[c]{@{}c@{}}RCL +\\ perceptual loss \end{tabular}} &
  20 &
  0.836 ± 0.125 &
  0.822 ± 0.131 &
  32.67 ± 29.14 &
  0.893 ± 0.115 &
  0.924 ± 0.089 \\ \hline
 &
   &
  100 &
  0.879 ± 0.115 &
  0.867 ± 0.121 &
  26.00 ± 30.40 &
  0.932 ± 0.091 &
  0.930 ± 0.099 \\ \cline{3-8} 
 &
   &
  50 &
  0.829 ± 0.170 &
  0.814 ± 0.175 &
  37.64 ± 45.12 &
  0.881 ± 0.171 &
  0.928 ± 0.098 \\ \cline{3-8} 
 &
  \multirow{-3}{*}{PIRL} &
  20 &
  0.825 ± 0.133 &
  0.810 ± 0.137 &
  36.89 ± 33.58 &
  0.867 ± 0.112 &
  0.934 ± 0.094 \\ \cline{2-8} 
 &
   &
  100 &
  0.892 ± 0.110 &
  0.878 ± 0.114 &
  20.89 ± 26.50 &
  0.930 ± 0.076 &
  \textbf{0.944 ± 0.093} \\ \cline{3-8} 
 &
   &
  50 &
  0.894 ± 0.101 &
  0.880 ± 0.106 &
  20.02 ± 24.85 &
  0.948 ± 0.053 &
  0.927 ± 0.102 \\ \cline{3-8} 
 &
  \multirow{-3}{*}{\begin{tabular}[c]{@{}c@{}}PIRL + \\ perceptual loss \end{tabular}} &
  20 &
  0.864 ± 0.117 &
  0.850 ± 0.124 &
  27.31 ± 27.91 &
  0.922 ± 0.089 &
  0.923 ± 0.089 \\ \cline{2-8} 
 &
   &
  100 &
  0.888 ± 0.105 &
  0.875 ± 0.111 &
  22.42 ± 25.56 &
  0.937 ± 0.072 &
  0.933 ± 0.099 \\ \cline{3-8} 
 &
   &
  50 &
  0.875 ± 0.135 &
  0.861 ± 0.138 &
  24.86 ± 33.02 &
  0.931 ± 0.092 &
  0.926 ± 0.097 \\ \cline{3-8} 
 &
  \multirow{-3}{*}{\begin{tabular}[c]{@{}c@{}}RCL \end{tabular}} &
  20 &
  \textbf{0.887 ± 0.100} &
  \textbf{0.876 ± 0.105} &
  \textbf{21.87 ± 25.15} &
  0.970 ± 0.045 &
  0.902 ± 0.102 \\ \cline{2-8} 
 &
   &
  100 &
  0.905 ± 0.103 &
  0.893 ± 0.109 &
  18.25 ± 25.71 &
  0.957 ± 0.057 &
  0.931 ± 0.105 \\ \cline{3-8} 
 &
   &
  50 &
  0.881 ± 0.113 &
  0.868 ± 0.118 &
  23.58 ± 27.03 &
  0.939 ± 0.072 &
  0.926 ± 0.101 \\ \cline{3-8} 
\multirow{-12}{*}{\begin{tabular}[c]{@{}c@{}}Cross-patch\\ Jigsaw\end{tabular}} &
  \multirow{-3}{*}{\begin{tabular}[c]{@{}c@{}}RCL + \\ perceptual loss \end{tabular}} &
  20 &
  0.874 ± 0.113 &
  0.861 ± 0.117 &
  26.98 ± 29.84 &
  0.926 ± 0.079 &
  0.930 ± 0.097 \\ \hline
 &
   &
  100 &
  0.876 ± 0.127 &
  0.863 ± 0.133 &
  25.18 ± 31.45 &
  0.915 ± 0.110 &
  0.943 ± 0.091 \\ \cline{3-8} 
 &
   &
  50 &
  0.892 ± 0.108 &
  0.880 ± 0.114 &
  21.39 ± 27.22 &
  \textbf{0.969 ± 0.052} &
  0.907 ± 0.104 \\ \cline{3-8} 
 &
  \multirow{-3}{*}{PIRL} &
  20 &
  0.849 ± 0.102 &
  0.834 ± 0.108 &
  28.10 ± 25.52 &
  0.935 ± 0.074 &
  0.891 ± 0.101 \\ \cline{2-8} 
 &
   &
  100 &
  0.910 ± 0.104 &
  0.900 ± 0.109 &
  17.19 ± 25.68 &
  0.966 ± 0.048 &
  0.930 ± 0.108 \\ \cline{3-8} 
 &
   &
  50 &
  0.899 ± 0.099 &
  \textbf{0.887 ± 0.105} &
  \textbf{19.26 ± 24.73} &
  \textbf{0.970 ± 0.047} &
  0.911 ± 0.103 \\ \cline{3-8} 
 &
  \multirow{-3}{*}{\begin{tabular}[c]{@{}c@{}}PIRL + \\ perceptual loss \end{tabular}} &
  20 &
  0.858 ± 0.111 &
  0.844 ± 0.114 &
  28.37 ± 28.15 &
  0.914 ± 0.084 &
  0.925 ± 0.095 \\ \cline{2-8} 
 &
   &
  100 &
  0.901 ± 0.114 &
  0.888 ± 0.118 &
  19.94 ± 27.82 &
  0.948 ± 0.067 &
  0.936 ± 0.100 \\ \cline{3-8} 
 &
   &
  50 &
  0.895 ± 0.110 &
  0.881 ± 0.115 &
  20.15 ± 27.06 &
  0.946 ± 0.065 &
  0.930 ± 0.100 \\ \cline{3-8} 
 &
  \multirow{-3}{*}{\begin{tabular}[c]{@{}c@{}} RCL \end{tabular}} &
  20 &
  0.817 ± 0.131 &
  0.804 ± 0.136 &
  38.16 ± 31.83 &
  0.865 ± 0.116 &
  0.932 ± 0.097 \\ \cline{2-8} 
 &
   &
  100 &
  \textbf{0.914 ± 0.098} &
  \textbf{0.902 ± 0.104} &
  \textbf{16.65 ± 24.66} &
  \textbf{0.972 ± 0.039} &
  0.926 ± 0.104 \\ \cline{3-8} 
 &
   &
  50 &
  \textbf{0.902 ± 0.109} &
  \textbf{0.890 ± 0.114} &
  \textbf{18.96 ± 26.70} &
  0.967 ± 0.051 &
  0.918 ± 0.101 \\ \cline{3-8} 
\multirow{-12}{*}{\begin{tabular}[c]{@{}c@{}}Cross-patch\\ Jigsaw \\+ Freq\end{tabular}} &
  \multirow{-3}{*}{\begin{tabular}[c]{@{}c@{}} RCL + \\ perceptual loss \end{tabular}} &
  20 &
  0.860 ± 0.118 &
  0.847 ± 0.124 &
  28.05 ± 28.83 &
  0.925 ± 0.088 &
  0.918 ± 0.094 \\ \hline

\end{tabular}
\end{table*}

\subsection{Quantitative Results on Generalisability}
Table \ref{tab:BUSI_BrEaST_optimised_results} shows the segmentation performance of our method variations on the held-out UDIAT dataset after training on combined BUSI and BrEaST datasets. All methods perform strongly when trained on the full BUSI and BrEaST datasets. The supervised (Res-UNet \cite{Zhengxin_2018_ResUNet}) baseline achieves $0.903$, $0.889$ and $19.19$ in DSC, JC and HD, respectively. Our best approach using all training data is Jig+Freq with PIRL+percep. This performs similarly to the Jig PIRL \cite{misra_self-supervised_2020} baseline with a performance of $0.927$, $0.915$ and $13.46$ in DSC, JC and HD, respectively. However, we observe a significant decrease in DSC performance by $23.6\%$ in the supervised (Res-UNet \cite{Zhengxin_2018_ResUNet}) approach from $100\%$ to $20\%$ training samples. This issue is alleviated with SSL approaches, as demonstrated in our generalisability study (Table \ref{tab:BUSI_BrEaST_optimised_results}), where the reduction in DSC performance from $100\%$ to $20\%$ training samples is smaller compared to the supervised (Res-UNet \cite{Zhengxin_2018_ResUNet}) baseline. The best methods with the smallest performance drop were Jig+Freq RCL+percep, with similar performance using $100\%$ or $50\%$ training samples, and Jig+Freq PIRL+percep, which dropped only by $3\%$ from $100\%$ to $20\%$ training samples.



Furthermore, using $50\%$ of training samples from BUSI and BrEaST datasets, we demonstrate improved performance with frequency augmentation added in the pretext task. The top two performing approaches are Jig+Freq using RCL+percep and CP-Jig+Freq using PIRL+percep. The best-performing approach is the CP-Jig+Freq task with PIRL+percep achieving a DSC score of $0.914$, a JC score of $0.902$ and an HD of $16.54$. This is a significant improvement of $1.8\%$, $1.9\%$ and $14.3\%$ compared to the Jigsaw PIRL baseline (p=0.039) \cite{misra_self-supervised_2020}. Using $20\%$ training samples, the top two performing approaches, both use the Jig+Freq pretext task and the best performance is achieved from the Jig+Freq PIRL+percep method. This method outperforms the PIRL (SSL) baseline \cite{misra_self-supervised_2020} by $1.5\%$, $1.4\%$ and $12.6\%$ in DSC, JC and HD, respectively (p = 0.090).

Overall, our generalisability study demonstrates improved segmentation performance in limited data scenarios using SSL approaches, outperforming the supervised (Res-UNet \cite{Zhengxin_2018_ResUNet}) baseline. Furthermore, frequency augmentation in our data-engineered pretext task, improves segmentation performance across all $100\%$, $50\%$ and $20\%$ training data conditions. Additionally, combining perceptual loss with PIRL and RCL shows improved performance in limited data scenarios. 

\begin{table*}[t!h!]
\centering
\caption{Assessing generalisability of methods on UDIAT dataset. BUSI and BrEaST datasets are used for pretext and downstream training. All downstream models use Res-UNet with ResNet50 encoder. Here, SD is the standard deviation.}
\label{tab:BUSI_BrEaST_optimised_results}
\begin{tabular}{c|c|c|l|l|l|l|l}
\hline
\textbf{\begin{tabular}[c]{@{}c@{}}Pretext \\ task\end{tabular}} &
  \textbf{Method} &
  \textbf{\begin{tabular}[c]{@{}c@{}}\% train \\ samples\end{tabular}} &
  \multicolumn{1}{c|}{\textbf{DSC} $\pm$ SD} &
  \multicolumn{1}{c|}{\textbf{JC} $\pm$ SD} &
  \multicolumn{1}{c|}{\textbf{HD} $\pm$ SD} &
  \multicolumn{1}{c|}{\textbf{PPV} $\pm$ SD} &
  \multicolumn{1}{c}{\textbf{Rec.} $\pm$ SD} \\ \hline
\multirow{3}{*}{N/A} &
  \multirow{3}{*}{\begin{tabular}[c]{@{}c@{}}Res-UNet~\cite{Zhengxin_2018_ResUNet}\\ (Supervised)\end{tabular}} &
  100 &
  0.903 ± 0.100 &
  0.889 ± 0.103 &
  19.19 ± 26.22 &
  0.943 ± 0.091 &
  0.942 ± 0.073 \\ \cline{3-8} 
 &
   &
  50 &
  0.778 ± 0.164 &
  0.761 ± 0.165 &
  49.79 ± 43.35 &
  0.801 ± 0.161 &
  0.952 ± 0.081 \\ \cline{3-8} 
&
&
  20 &
  0.690 ± 0.204 &
  0.670 ± 0.208 &
  79.92 ± 62.64 &
  0.702 ± 0.219 &
  \textbf{0.962 ± 0.058} \\ \hline
\multirow{12}{*}{Jigsaw} &
  \multirow{3}{*}{\begin{tabular}[c]{@{}c@{}}PIRL \cite{misra_self-supervised_2020}\\ Baseline\end{tabular}} &
  100 &
  \textbf{0.928 ± 0.080} &
  \textbf{0.916 ± 0.085} &
  \textbf{12.76 ± 16.92} &
  0.957 ± 0.054 &
  0.955 ± 0.074 \\ \cline{3-8} 
 &
   &
  50 &
  0.898 ± 0.087 &
  0.885 ± 0.089 &
  19.29 ± 20.49 &
  0.946 ± 0.059 &
  0.935 ± 0.085 \\ \cline{3-8} 
 &
   &
  20 &
  0.883 ± 0.115 &
  0.872 ± 0.114 &
  23.72 ± 29.37 &
  0.957 ± 0.070 &
  0.909 ± 0.109 \\ \cline{2-8} 
 &
  \multirow{3}{*}{\begin{tabular}[c]{@{}c@{}}PIRL + \\ perceptual loss \end{tabular}} &
  100 &
  0.910 ± 0.101 &
  0.898 ± 0.102 &
  17.68 ± 24.40 &
  0.960 ± 0.069 &
  0.936 ± 0.081 \\ \cline{3-8} 
 &
   &
  50 &
  0.871 ± 0.130 &
  0.860 ± 0.131 &
  27.92 ± 33.85 &
  0.930 ± 0.107 &
  0.926 ± 0.095 \\ \cline{3-8} 
 &
   &
  20 &
  0.869 ± 0.123 &
  0.859 ± 0.122 &
  28.62 ± 32.10 &
  0.957 ± 0.071 &
  0.901 ± 0.122 \\ \cline{2-8} 
 &
  \multirow{3}{*}{\begin{tabular}[c]{@{}c@{}} RCL \end{tabular}} &
  100 &
  0.922 ± 0.082 &
  0.909 ± 0.084 &
  14.41 ± 20.01 &
  0.961 ± 0.059 &
  0.945 ± 0.065 \\ \cline{3-8} 
 &
   &
  50 &
  0.896 ± 0.103 &
  0.883 ± 0.104 &
  21.08 ± 25.50 &
  0.938 ± 0.086 &
  0.942 ± 0.069 \\ \cline{3-8} 
 &
   &
  20 &
  0.854 ± 0.123 &
  0.844 ± 0.124 &
  30.54 ± 31.25 &
  0.959 ± 0.083 &
  0.881 ± 0.115 \\ \cline{2-8} 
 &
  \multirow{3}{*}{\begin{tabular}[c]{@{}c@{}}RCL + \\ perceptual loss \end{tabular}} &
  100 &
  0.902 ± 0.094 &
  0.887 ± 0.098 &
  19.24 ± 22.44 &
  0.920 ± 0.086 &
  0.965 ± 0.062 \\ \cline{3-8} 
 &
   &
  50 &
  0.906 ± 0.092 &
  0.891 ± 0.096 &
  19.40 ± 23.17 &
  0.916 ± 0.082 &
  \textbf{0.970 ± 0.058} \\ \cline{3-8} 
 &
   &
  20 &
  0.891 ± 0.111 &
  0.879 ± 0.111 &
  22.93 ± 28.08 &
  0.944 ± 0.077 &
  0.932 ± 0.098 \\ \hline
\multirow{12}{*}{\begin{tabular}[c]{@{}c@{}} Jigsaw\\ + Freq\end{tabular}} &
  \multirow{3}{*}{PIRL} &
  100 &
  0.922 ± 0.090 &
  0.910 ± 0.092 &
  13.90 ± 19.38 &
  \textbf{0.972 ± 0.057} &
  0.936 ± 0.083 \\ \cline{3-8} 
 &
   &
  50 &
  0.892 ± 0.112 &
  0.881 ± 0.113 &
  22.18 ± 27.17 &
  0.949 ± 0.070 &
  0.928 ± 0.107 \\ \cline{3-8} 
 &
   &
  20 &
  \textbf{0.892 ± 0.113} &
  \textbf{0.883 ± 0.111} &
  \textbf{21.76 ± 27.81} &
  0.981 ± 0.035 &
  0.899 ± 0.113 \\ \cline{2-8} 
 &
  \multirow{3}{*}{\begin{tabular}[c]{@{}c@{}}PIRL + \\ perceptual loss \end{tabular}} &
  100 &
  \textbf{0.927 ± 0.080} &
  \textbf{0.915 ± 0.082} &
  13.46 ± 19.13 &
  0.961 ± 0.047 &
  0.952 ± 0.083 \\ \cline{3-8} 
 &
   &
  50 &
  0.895 ± 0.113 &
  0.885 ± 0.114 &
  21.38 ± 26.87 &
  \textbf{0.964 ± 0.070} &
  0.918 ± 0.103 \\ \cline{3-8} 
 &
   &
  20 &
  \textbf{0.896 ± 0.104} &
  \textbf{0.884 ± 0.104} &
  \textbf{20.74 ± 25.62} &
  0.937 ± 0.074 &
  \textbf{0.944 ± 0.095} \\ \cline{2-8} 
 &
  \multirow{3}{*}{\begin{tabular}[c]{@{}c@{}} RCL \end{tabular}} &
  100 &
  0.810 ± 0.191 &
  0.794 ± 0.193 &
  45.31 ± 54.31 &
  0.824 ± 0.195 &
  0.965 ± 0.054 \\ \cline{3-8} 
 &
   &
  50 &
  0.844 ± 0.116 &
  0.829 ± 0.116 &
  32.58 ± 29.30 &
  0.898 ± 0.112 &
  0.927 ± 0.080 \\ \cline{3-8} 
 &
   &
  20 &
  0.791 ± 0.139 &
  0.778 ± 0.138 &
  45.02 ± 34.92 &
  0.865 ± 0.143 &
  0.907 ± 0.099 \\ \cline{2-8} 
 &
  \multirow{3}{*}{\begin{tabular}[c]{@{}c@{}} RCL + \\ perceptual loss \end{tabular}} &
  100 &
  0.904 ± 0.109 &
  0.892 ± 0.111 &
  18.25 ± 24.42 &
  \textbf{0.967 ± 0.066} &
  0.923 ± 0.096 \\ \cline{3-8} 
 &
   &
  50 &
  \textbf{0.907 ± 0.097} &
  \textbf{0.893 ± 0.100} &
  \textbf{17.63 ± 20.98} &
  0.930 ± 0.078 &
  0.959 ± 0.069 \\ \cline{3-8} 
 &
   &
  20 &
  0.848 ± 0.149 &
  0.834 ± 0.151 &
  34.87 ± 40.33 &
  0.892 ± 0.141 &
  0.935 ± 0.097 \\ \hline
\multirow{12}{*}{\begin{tabular}[c]{@{}c@{}}Cross-patch\\ Jigsaw\end{tabular}} &
  \multirow{3}{*}{PIRL} &
  100 &
  0.901 ± 0.120 &
  0.888 ± 0.121 &
  21.25 ± 34.13 &
  0.937 ± 0.106 &
  0.949 ± 0.072 \\ \cline{3-8} 
 &
   &
  50 &
  0.893 ± 0.106 &
  0.880 ± 0.107 &
  21.28 ± 25.64 &
  0.942 ± 0.072 &
  0.933 ± 0.098 \\ \cline{3-8} 
 &
   &
  20 &
  0.854 ± 0.126 &
  0.847 ± 0.125 &
  32.47 ± 33.63 &
  \textbf{0.990 ± 0.036} &
  0.856 ± 0.126 \\ \cline{2-8} 
 &
  \multirow{3}{*}{\begin{tabular}[c]{@{}c@{}}PIRL + \\ perceptual loss \end{tabular}} &
  100 &
  0.906 ± 0.107 &
  0.893 ± 0.109 &
  19.25 ± 26.01 &
  0.933 ± 0.087 &
  0.958 ± 0.087 \\ \cline{3-8} 
 &
   &
  50 &
  0.895 ± 0.103 &
  0.884 ± 0.103 &
  21.26 ± 24.65 &
  \textbf{0.961 ± 0.074} &
  0.921 ± 0.094 \\ \cline{3-8} 
 &
   &
  20 &
  0.850 ± 0.149 &
  0.842 ± 0.149 &
  33.22 ± 39.00 &
  0.947 ± 0.102 &
  0.892 ± 0.134 \\ \cline{2-8} 
 &
  \multirow{3}{*}{\begin{tabular}[c]{@{}c@{}} RCL \end{tabular}} &
  100 &
  0.891 ± 0.116 &
  0.877 ± 0.119 &
  22.06 ± 27.96 &
  0.903 ± 0.110 &
  \textbf{0.967 ± 0.061} \\ \cline{3-8} 
 &
   &
  50 &
  0.859 ± 0.134 &
  0.846 ± 0.134 &
  30.98 ± 37.45 &
  0.914 ± 0.117 &
  0.926 ± 0.094 \\ \cline{3-8} 
 &
   &
  20 &
  0.869 ± 0.111 &
  0.860 ± 0.112 &
  26.56 ± 26.81 &
  0.968 ± 0.060 &
  0.889 ± 0.110 \\ \cline{2-8} 
 &
  \multirow{3}{*}{\begin{tabular}[c]{@{}c@{}} RCL + \\ perceptual loss \end{tabular}} &
  100 &
  0.923 ± 0.084 &
  0.911 ± 0.089 &
  14.97 ± 22.67 &
  0.959 ± 0.068 &
  0.949 ± 0.073 \\ \cline{3-8} 
 &
   &
  50 &
  0.891 ± 0.106 &
  0.876 ± 0.110 &
  23.91 ± 27.72 &
  0.905 ± 0.104 &
  \textbf{0.966 ± 0.050} \\ \cline{3-8} 
 &
   &
  20 &
  0.885 ± 0.112 &
  0.875 ± 0.113 &
  23.48 ± 27.82 &
  0.961 ± 0.064 &
  0.912 ± 0.114 \\ \hline
\multirow{12}{*}{\begin{tabular}[c]{@{}c@{}}Cross-patch\\ Jigsaw \\+ Freq\end{tabular}} &
  \multirow{3}{*}{PIRL} &
  100 &
  0.921 ± 0.092 &
  0.909 ± 0.094 &
  14.36 ± 20.48 &
  0.969 ± 0.043 &
  0.938 ± 0.094 \\ \cline{3-8} 
 &
   &
  50 &
  0.885 ± 0.109 &
  0.869 ± 0.111 &
  24.63 ± 28.14 &
  0.900 ± 0.099 &
  0.962 ± 0.068 \\ \cline{3-8} 
 &
   &
  20 &
  0.876 ± 0.126 &
  0.867 ± 0.126 &
  26.03 ± 32.09 &
  0.966 ± 0.072 &
  0.899 ± 0.118 \\ \cline{2-8} 
 &
  \multirow{3}{*}{\begin{tabular}[c]{@{}c@{}}PIRL + \\ perceptual loss \end{tabular}} &
  100 &
  0.926 ± 0.088 &
  0.913 ± 0.091 &
  \textbf{13.08 ± 18.67} &
  0.945 ± 0.079 &
  \textbf{0.966 ± 0.057} \\ \cline{3-8} 
 &
   &
  50 &
  \textbf{0.914 ± 0.095} &
  \textbf{0.902 ± 0.099} &
  \textbf{16.54 ± 21.87} &
  0.948 ± 0.080 &
  0.950 ± 0.075 \\ \cline{3-8} 
 &
   &
  20 &
  0.865 ± 0.126 &
  0.858 ± 0.125 &
  29.39 ± 33.08 &
  0.986 ± 0.029 &
  0.870 ± 0.131 \\ \cline{2-8} 
 &
  \multirow{3}{*}{\begin{tabular}[c]{@{}c@{}} RCL \end{tabular}} &
  100 &
  0.877 ± 0.117 &
  0.862 ± 0.118 &
  25.50 ± 31.05 &
  0.923 ± 0.102 &
  0.935 ± 0.075 \\ \cline{3-8} 
 &
   &
  50 &
  0.870 ± 0.128 &
  0.859 ± 0.128 &
  26.82 ± 33.04 &
  0.939 ± 0.113 &
  0.916 ± 0.083 \\ \cline{3-8} 
 &
   &
  20 &
  0.842 ± 0.120 &
  0.829 ± 0.121 &
  32.47 ± 29.44 &
  0.922 ± 0.109 &
  0.902 ± 0.088 \\ \cline{2-8} 
 &
  \multirow{3}{*}{\begin{tabular}[c]{@{}c@{}} RCL +\\ perceptual loss \end{tabular}} &
  100 &
  0.902 ± 0.106 &
  0.889 ± 0.109 &
  19.01 ± 25.21 &
  0.930 ± 0.090 &
  0.955 ± 0.077 \\ \cline{3-8} 
 &
   &
  50 &
  0.899 ± 0.097 &
  0.884 ± 0.101 &
  19.78 ± 22.47 &
  0.926 ± 0.085 &
  0.954 ± 0.078 \\ \cline{3-8} 
 &
   &
  20 &
  0.857 ± 0.129 &
  0.850 ± 0.128 &
  31.65 ± 34.85 &
  \textbf{0.987 ± 0.025} &
  0.861 ± 0.134 \\ \hline

\end{tabular}
\end{table*}

\subsection{Qualitative Analysis}
\begin{figure*}[]
    \centering    \includegraphics[width=1\textwidth]{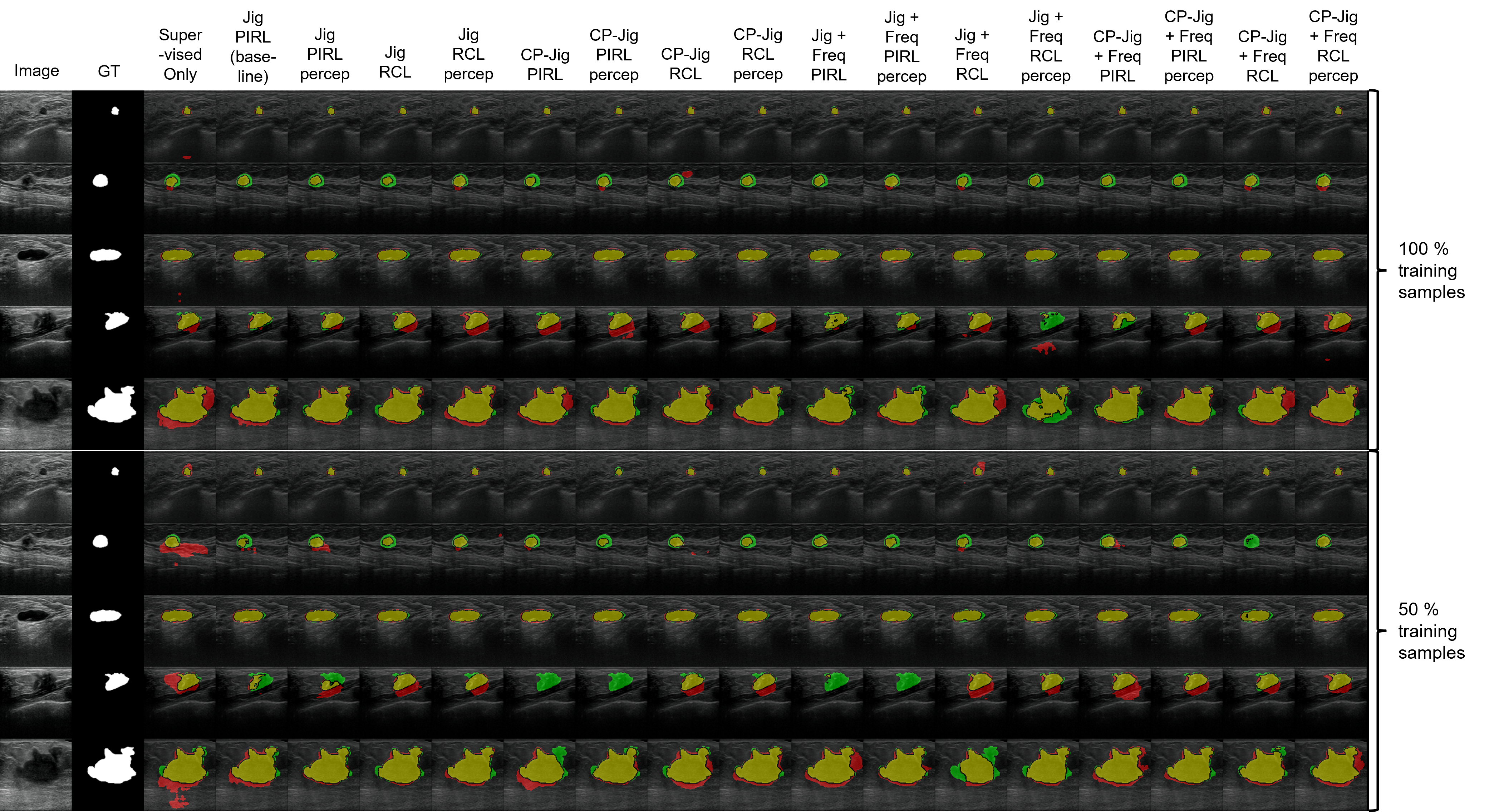}
    \caption{Qualitative evaluation of generalisability study on held out UDIAT dataset. 5 examples were chosen from small regular-shaped lesions to larger irregular-shaped lesions. Segmentation predictions across all 17 methods are presented, using 100\% and 50\% training samples. Yellow indicates ground truth labels overlayed onto the original image. Green indicates areas of under-segmentation and red indicates areas of over-segmentation relative to the ground truth.}
    \label{fig:Qual_Generalisability_examples}
\end{figure*}

Segmentation predictions from small regular to large irregular tumour shapes for each approach is shown in Fig. \ref{fig:Qual_Generalisability_examples} for our generalisability study.
We can observe that all methods effectively segment smooth and regular shaped lesions, whilst irregular lesions are the most challenging. The supervised (Res-UNet \cite{Zhengxin_2018_ResUNet} baseline is prone to over-segmentation (indicated in red in Fig. \ref{fig:Qual_Generalisability_examples}), particularly when using $50\%$ of training samples. Generally, the SSL approaches provide improved segmentation. Across all methods, we observe worse performance in segmenting the tumour shown in row 4 when using $100\%$ of the training samples and in row $9$ when using $50\%$. Using $100\%$ of the training samples, over-segmentation of this tumour is common across all methods except Jig+Freq RCL+percep. Similarly, over-segmentation remains a common issue on $50\%$ of the training samples. Under this condition, CP-Jig PIRL, CP-Jig PIRL+percep, Jig+Freq PIRL, and Jig+Freq PIRL+percep completely fail to segment this tumour, while the Jig PIRL baseline and Jig PIRL+percep under-segment it.

Across all tumour examples in both the $100\%$ and $50\%$ training scenarios, CP-Jig+Freq with RCL+percep performs best overall. It captures irregular boundary shapes most accurately compared to the ground truth mask, particularly in the tumour shown in rows $4$ and $9$. CP-Jig RCL, CP-Jig RCL+percep and CP-Jig+Freq RCL also segment well, capturing the full tumour area without excessive over/under segmentation in most examples. PIRL based methods (incl. Jig PIRL baseline, Jig PIRL+percep, CP-Jig PIRL, CP-Jig PIRL+percep, Jig+Freq PIRL, Jig+Freq PIRL+percep) often perform worse, particularly when segmenting more irregular shaped tumours and more commonly under-segment the complete shape of the tumour area (see row $9$ in Fig. \ref{fig:Qual_Generalisability_examples}) CP-Jig+Freq PIRL performs slightly better but tends to over-segment, while CP-Jig+Freq PIRL+percep achieves the best performance among PIRL methods. This is also reflected in Table \ref{tab:BUSI_BrEaST_optimised_results}, where it achieves top two performance on $50\%$ of the training samples. However, qualitatively, the RCL counterpart (CP-Jig+Freq RCL+percep) segments smaller irregular tumour areas more effectively (see Fig. \ref{fig:Qual_Generalisability_examples}, rows 4 and 9).



\subsection{Ablation Study}
\label{sec:Ablation_study}
We include an ablation study to investigate the $\lambda$ weighting for methods with combined loss functions (i.e., PIRL+percep and RCL+percep) in Table~\ref{tab:ablation_lambda}. Ablation results are demonstrated on the BUSI dataset using the validation set. We also use the baseline pretext learning Jigsaw task for all ablation results. Our ablation results demonstrate a $\lambda$ value of $0.75$ and $0.1$ are optimal weights in PIRL+percep and RCL+percep, respectively. 

\begin{table}[t!]
\centering
\caption{Effect of varying $\lambda$ on combined loss methods (Eq.\ref{eqn:Combined_loss}).}
\label{tab:ablation_lambda}
\begin{tabular}{c|c|c}
\hline
\textbf{Method} & \textbf{$\lambda$ setting} & \textbf{DSC} \\ \hline
\multirow{4}{*}{\begin{tabular}[c]{@{}c@{}}PIRL + \\ perceptual loss \end{tabular}} & 0.1 & 0.813 \\ \cline{2-3} 
 & 0.25 & 0.883
 \\ \cline{2-3} 
 & 0.5 & 0.866 \\ \cline{2-3} 
 & 0.75 & \textbf{0.888} \\ \hline
\multirow{4}{*}{\begin{tabular}[c]{@{}c@{}}RCL +  \\ perceptual loss \end{tabular}} & 0.1 & \textbf{0.890} \\ \cline{2-3} 
 & 0.25 & 0.885 \\ \cline{2-3} 
 & 0.5 & 0.881 \\ \cline{2-3} 
 & 0.75 & 0.873 \\ \hline
\end{tabular}%
\end{table}

\section{Discussion}
Supervised learning models have shown promising improvements in US image segmentation \cite{wu_cross-image_2023, zhang_hau-net_2024, jiang_hybrid_2023}. However, performance is significantly affected when training samples are limited \cite{vanberlo_survey_2024}. Our results in all datasets demonstrate that the supervised baseline is consistently among the lowest performing methods when either $50\%$ or $20\%$ training samples are used. This finding suggests that our SSL approaches can effectively learn meaningful representations for US data during pretext learning that supports the downstream segmentation task (Table~\ref{tab:BUSI_optimised_results}--\ref{tab:BUSI_BrEaST_optimised_results}). 
Furthermore, supervised model generalisability is significantly impacted when training samples are limited. This is evident in our generalisability study (see Table \ref{tab:BUSI_BrEaST_optimised_results}). The baseline supervised method shows a significant $23.6\%$ drop in DSC and a $76\%$ increase in HD when training data is reduced from $100\%$ to $20\%$. This highlights the benefits of SSL in improving segmentation performance on out-of-distribution data, particularly when downstream data is limited. Therefore, incorporating SSL techniques into US model development can enhance clinical applicability. Improving performance on data acquired using different hardware systems, patient populations, and clinical operators relative to training data improves model robustness and real-world method adoption. 

To enhance the performance of contrastive SSL for segmentation of US images, we first introduced a data-engineered, domain-inspired pretext task aimed at US B-mode image data. Our results demonstrate that our frequency augmentation in pretext learning benefits the downstream segmentation task performance in US images. This addition occurs in the top two approaches across all datasets explored, including our generalisability study and in all training proportions of $100\%$, $50\%$ and $20\%$. Our frequency augmentation improves feature learning by exposing the model to a broader range of image degradations. This is reflective of real-world clinical settings, where image quality can vary significantly across different hardware, users and patients. Our proposed Cross-patch Jigsaw spatial transformation performed comparably well to the Jigsaw pretext task. It occurred within the top two approaches (often with frequency augmentation included) when $50\%$ of the training data is used across all datasets explored, including the generalisability study. This is encouraging, as this transformation was designed to preserve partial layer-wise structural information within the ultrasound image. In more complex imaging domains, such as abdominal ultrasound, which involve additional structures and more detailed layer-wise information, this approach could offer an improved alternative. 

In addition to our novel data engineered pretext task, we proposed a combined loss function, which includes RCL and perceptual loss. From our results perceptual loss benefits both PIRL and RCL, with perceptual loss occurring in the top two performing methods across all training data proportions and datasets explored, occurring in $11$ of $12$ cases. This loss component was added to focus representation learning on higher level more abstract features from deep within the encoder network (in our case layer 40 from ResNet50). Our results indicate that the use of perceptual loss is beneficial for feature representation learning in ultrasound for downstream segmentation tasks, particularly in combination with our relation contrastive loss (RCL). RCL consistently ranks among the top two performers in each data set, demonstrating notable benefits in limited data scenarios. Specifically, RCL achieves top-two segmentation performance using $50\%$ and $20\%$ of training samples in the BUSI, BrEaST and UDIAT datasets. This shows that by combining encoder-learned features with a relation network, we can enhance learned feature representations by improving the differentiation between positive and negative samples during pretext learning. Furthermore, our qualitative results (see Fig. \ref{fig:Qual_Generalisability_examples}) show that our approach, RCL with perceptual loss, along with our proposed spatial and frequency pretext tasks, can improve the segmentation of irregular tumours.

\section{Conclusion}
To the best of our knowledge, this is the most comprehensive exploration of domain-inspired data-engineered pretext tasks for US image analysis within a contrastive self-supervised learning framework. We introduced the novel Relation Contrastive Loss (RCL) combined with perceptual loss, a unique and previously unexplored approach. This method demonstrated improved downstream segmentation performance, particularly in limited data scenarios. Furthermore,  combining contrastive loss with perceptual loss consistently improves segmentation performance and generalisability in noisy US image data. Additionally, we showed that frequency-based band-stop filtering as an augmentation technique in contrastive self-supervised learning improves the learning of generalisable features. Our novel Cross-Patch Jigsaw approach delivered comparable performance to the traditional Jigsaw task, and combining our spatial and frequency-based pretext tasks with RCL and perceptual loss further improved segmentation of irregular tumour areas. In our future work, we aim to apply and extend this method to US images in more challenging clinical domains, such as abdominal ultrasound for bowel and gall bladder segmentation. 


\bibliographystyle{IEEEtran}
\bibliography{references}

\end{document}